\documentclass[journal]{IEEEtran}

\usepackage[pdftex]{graphicx}
\usepackage{cite}
\usepackage{amsopn}
\usepackage{booktabs} 
\usepackage{url}
\usepackage{color}
\usepackage{multirow}
\usepackage{xcolor}
\usepackage[switch]{lineno}
\usepackage{graphicx}
\usepackage{subfigure}
\usepackage{url}
\usepackage{epstopdf}
\usepackage{soul}
\usepackage{xcolor}
\usepackage{amsmath}
\usepackage{textcomp}
\usepackage{verbatim}
\usepackage{bm}
\usepackage{hyperref}

\begin{document}

\title{Dynamic Spatial-Temporal Representation Learning for {\color{black}Traffic} Flow Prediction}

\author{Lingbo Liu,
        Jiajie Zhen,
        Guanbin Li,
        Geng Zhan,
        Zhaocheng He,
        Bowen Du,
        and~Liang Lin
\thanks{L. Liu, J. Zhen, G. Li and L. Lin are with the School of Data and Computer Science, Sun Yat-Sen University, China, 510000 (e-mail: liulingb@mail2.sysu.edu.cn; qiuzhl3@mail2.sysu.edu.cn; liguanbin@mail.sysu.edu.cn; linliang@ieee.org).}
\thanks{G. Zhan is with the School of Electrical and Information Engineering, the University of Sydney, Australia, 2000 (e-mail: gengzh0308@gmail.com).}
\thanks{Z. He is with the School of Intelligent Systems Engineering, Sun Yat-Sen University, China, 510000 (e-mail: hezhch@mail.sysu.edu.cn).}
\thanks{B. Du is with the State Key Laboratory of Software Development Environment, Beihang University, China, 100191 (e-mail: dubowen@buaa.edu.cn).}
}

\markboth{IEEE Transactions on Intelligent Transportation Systems}
{Liu \MakeLowercase{\textit{et al.}}: ATFM}

\maketitle


\begin{abstract}
As a crucial component in intelligent transportation systems, traffic flow prediction has recently attracted widespread research interest in the field of artificial intelligence (AI) with the increasing availability of massive traffic mobility data. Its key challenge lies in how to integrate diverse factors (such as temporal rules and spatial dependencies) to infer the evolution trend of traffic flow.
To address this problem, we propose a unified neural network called Attentive Traffic Flow Machine (ATFM), which can effectively learn the spatial-temporal feature representations of traffic flow with an attention mechanism. In particular, our ATFM is composed of two progressive Convolutional Long Short-Term Memory (ConvLSTM~\cite{xingjian2015convolutional}) units connected with a convolutional layer. Specifically, the first ConvLSTM unit takes normal traffic flow features as input and generates a hidden state at each time-step, which is further fed into the connected convolutional layer for spatial attention map inference. The second ConvLSTM unit aims at learning the dynamic spatial-temporal representations from the attentionally weighted traffic flow features.
Further, we develop two deep learning frameworks based on ATFM to predict citywide short-term/long-term traffic flow by adaptively incorporating the sequential and periodic data as well as other external influences.
Extensive experiments on two standard benchmarks well demonstrate the superiority of the proposed method for traffic flow prediction. Moreover, to verify the generalization of our method, we also apply the customized framework to forecast the passenger pickup/dropoff demands in traffic prediction and show its superior performance.Our code and data are available at {\color{blue}\url{https://github.com/liulingbo918/ATFM}}.
%
\end{abstract}
\begin{IEEEkeywords}
traffic flow prediction, mobility data, spatial-temporal modeling, attentional recurrent neural network.
\end{IEEEkeywords}

\IEEEpeerreviewmaketitle
\section{Introduction}~\label{sec:introduction}
\IEEEPARstart{C}ity is the keystone of modern human living and individuals constantly migrate from rural areas to urban areas with urbanization. For instance, Delhi, the largest city in India, has a total of 29.4 million residents\footnote{\url{http://worldpopulationreview.com/world-cities/}}. Such a huge population brings a great challenge to urban management, especially in traffic control~\cite{zheng2014urban}.
To address this challenge, intelligent transportation systems (ITS)~\cite{zhang2011data} have been exhaustively studied for decades and have emerged as an efficient way of improving the efficiency of urban transportation. As a crucial component in ITS, traffic flow prediction~\cite{huang2014deep,lv2014traffic,polson2017deep} has recently attracted widespread research interest in both academic and industry communities, due to its huge potentials in many real-world applications (e.g., intelligent traffic diversion and travel optimization).

\begin{figure}
\centerline{
\includegraphics[width=0.600\columnwidth]{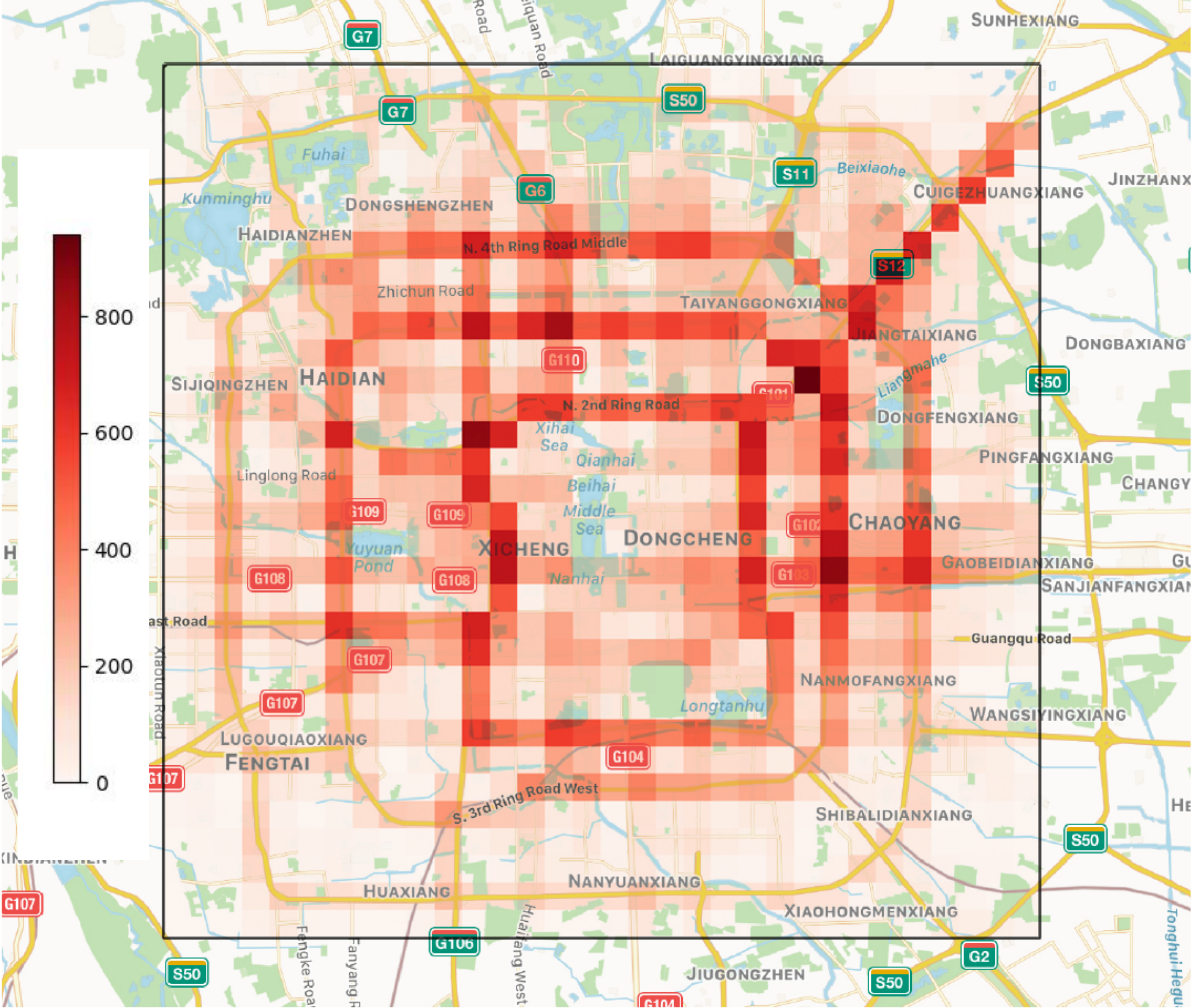}
\includegraphics[width=0.382\columnwidth]{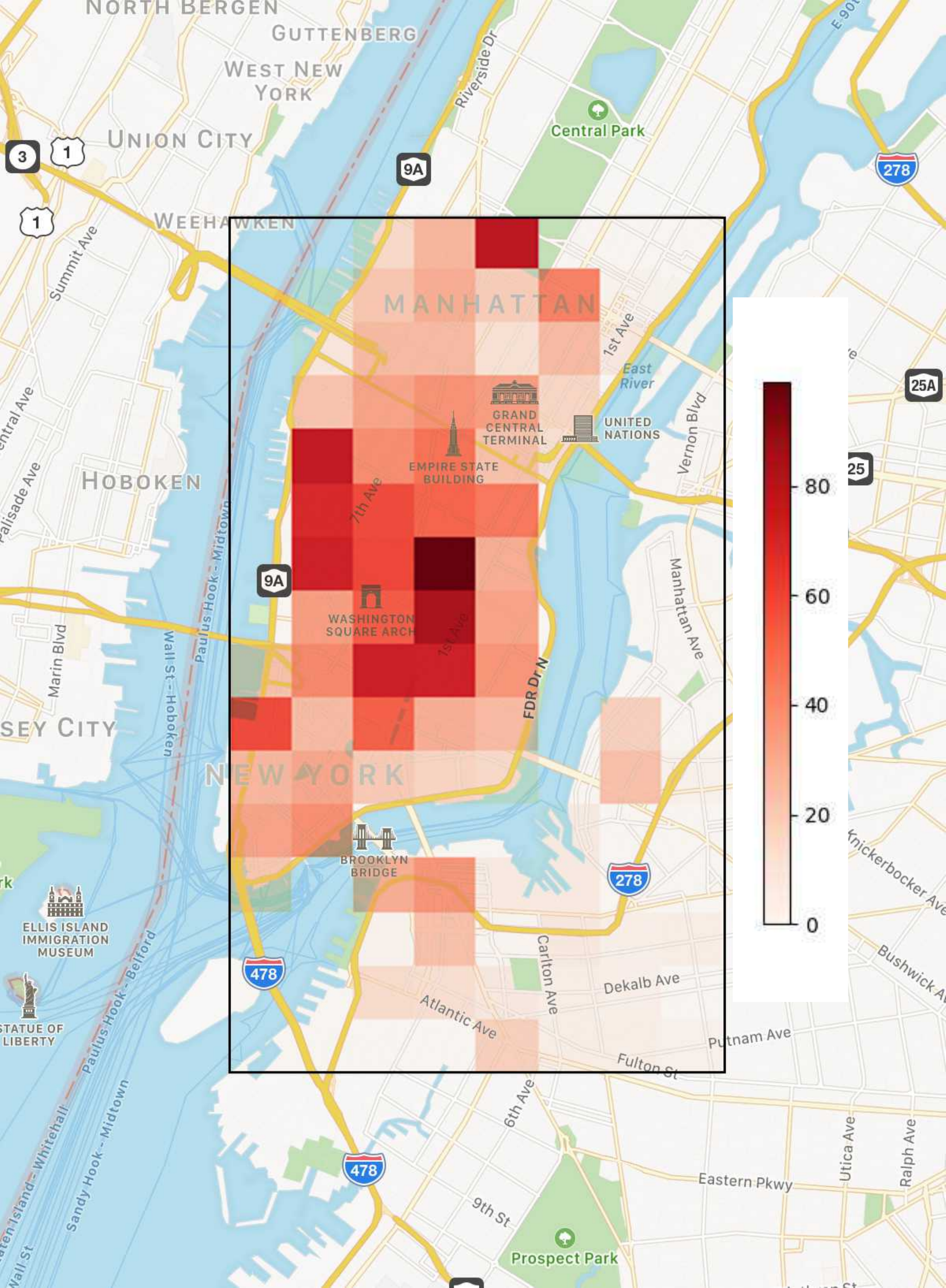}
}
\vspace{-2mm}
   \caption{Visualization of two traffic flow maps in Beijing and New York City. Following previous work~\cite{zhang2017deep}, we partition a city into a grid map based on the longitude and latitude and generate the historical traffic flow maps by measuring the number of taxicabs/bikes in each region with mobility data. The weight of a specific grid indicates the traffic density of its corresponding region during a time interval. In this work, we take these historical maps as input to forecast the future traffic flow.
   }
\vspace{0mm}
\label{fig:citywide_crow_flow_visual}
\end{figure}


In this paper, we aim to forecast the future traffic flow in a city with historical mobility data of taxicabs/bikes. Nowadays, we live in an era where ubiquitous digital devices are able to broadcast rich information about taxicabs/bikes mobility in real-time and at a high rate, which has rapidly increased the availability of large-scale mobility data (e.g., GPS signals or mobile phone signals). How to utilize these mobility data to predict traffic flow is still an open problem. In literature, numerous methods applied time series models (e.g., Auto-Regressive Integrated Moving Average (ARIMA)~\cite{shekhar2007adaptive} and Kalman filtering~\cite{guo2014adaptive}) to predict traffic flow at each individual location separately. Subsequently, some studies incorporated spatial information to conduct prediction~\cite{zheng2013time,deng2016latent}. However, these traditional models can not well capture the complex spatial-temporal dependency of traffic flow and this task is still far from being well solved in complex traffic systems.

Recently, notable successes have been achieved for citywide traffic flow prediction based on deep neural networks coupled with certain spatial-temporal priors~\cite{zhang2016dnn,zhang2017deep,xu2018predcnn,zhang2019flow}. In these works, the studied city is partitioned into a grid map based on the longitude and latitude, as shown in Fig.~\ref{fig:citywide_crow_flow_visual}. The historical traffic flow {\color{black}maps/tensors} generated from mobility data are fed into convolutional neural networks to forecast the future traffic flow. Nevertheless, there still exist several challenges limiting the performance of traffic flow analysis in complex scenarios.
{\bf{First}}, traffic flow data can vary greatly in temporal sequences and capturing such dynamic variations is non-trivial. {\color{black}However, previous methods~\cite{zhang2017deep,zhang2019flow,zhao2017lstm} simply applied convolutional operations or LSTM units to directly map historical data to future flow prediction which are not able to well model the temporal patterns.} {\bf{Second}}, the spatial dependencies between locations are not strictly stationary and the relation significance of a specific region may change from time to time. {\color{black}Unfortunately, most of the existing methods do not consider such dynamic spatial dependence of traffic flow.
} {\bf{Third}}, some internal periodic laws (e.g., traffic flow suddenly changing due to rush hours) and external factors (e.g., a precipitate rain) can greatly affect the situation of traffic flow, which increases the difficulty in learning traffic flow representations from data. {\color{black}Conventional works~\cite{zhang2017deep,zhang2018predicting} statically fuse these internal and external factors, which fail to flexibly generate effective representations to capture complex traffic flow patterns.}

To solve all above issues, we propose a novel spatial-temporal neural network, called Attentive Traffic Flow Machine~(ATFM), to adaptively exploit diverse factors that affect traffic flow evolution and at the same time produce the traffic flow estimation map in an end-to-end manner. The attention mechanism embedded in ATFM is designed to automatically discover the regions with primary impacts on the future flow prediction and simultaneously adjust the impacts of the different regions with different weights at each time-step. Specifically, our ATFM comprises two progressive ConvLSTM~\cite{xingjian2015convolutional} units.
The first one takes input from i) the original traffic flow features at each moment and ii) the memorized representations of previous moments, to compute the attentional weights. The second LSTM dynamically adjusts the spatial dependencies with the computed attentional map and generates superior spatial-temporal feature representation.
The proposed ATFM has the three following appealing properties. {\bf{First}}, it can effectively incorporate spatial-temporal information in feature representation and can flexibly compose solutions for traffic flow prediction with different types of input data. {\bf{Second}}, by integrating the deep attention mechanism~\cite{sharma2015action,lu2016knowing,liu2018crowd,tay2019aanet}, ATFM adaptively learns to represent the weights of each spatial location at each time-step, which allows the model to dynamically perceive the impact of the given area at a given moment for the future traffic flow. {\bf{Third}}, as a general and differentiable module, our ATFM can be effectively incorporated into various network architectures for end-to-end training.

Based on the proposed ATFM, we further develop a deep architecture for forecasting the citywide short-term traffic flow. Specifically, this customized framework consists of four components: {\bf{i)}} a normal feature extraction module, {\bf{ii)}} a sequential representation learning module, {\bf{iii)}} a periodic representation learning module and {\bf{iv)}} a temporally-varying fusion module. The middle two components are implemented by two parallel ATFMs for contextual dependencies modeling at different temporal scales, while the temporally-varying fusion module is proposed to adaptively merge the two separate temporal representations for traffic flow prediction. Finally, we extend and improve this framework to predict long-term traffic flow with an extra LSTM prediction network.
{\color{black}Notice that our framework is general. Besides citywide traffic flow prediction, it can also be applied to extensive traffic tasks (e.g., citywide passenger demand prediction, crowd flow prediction), if the following preprocessing procedures are satisfied: {\bf{i)}} the studied city is divided into a regular grid map and the raw traffic data is transformed into tensors, which is the most common form of structured data to fit the deep neural networks; {\bf{ii)}} the sequential data and periodic data have been recorded; {\bf{iii)}} the external factors (e.g., holiday information and meteorology information) are available, or else this submodule can be directly ignored.}

In summary, the contributions of this work are three-fold:
\begin{itemize}
\item We propose a novel neural network module called Attentive Traffic Flow Machine (ATFM), which incorporates two ConvLSTM units with an attention mechanism to infer the evolution trend of traffic flow via dynamic spatial-temporal feature representations learning.
\item We integrate the proposed ATFM in a customized deep framework for citywide traffic flow prediction, which effectively incorporates the sequential and periodic dependencies with a temporally-varying fusion module.
\item Extensive experiments on two public benchmarks of traffic flow prediction demonstrate the superiority of the proposed method.
\end{itemize}

A preliminary version of this work is published in~\cite{liu2018attentive}. In this work, we inherit the idea of dynamically learning the spatial-temporal representations and provide more details of the proposed method. Moreover, we extend this customized framework to forecast long-term traffic flow.
Further, we conduct a more comprehensive ablation study on our method and present more comparisons with state-of-the-art models under different settings (e.g., weekday, weekend, day and night).
Finally, we apply the proposed method to forecast the passenger pickup/dropoff demands and show that our method is generalizable to various traffic prediction tasks.

The rest of this paper is organized as follows. First, we review some related works of traffic flow analysis in Section~\ref{sec:review} and provide some preliminaries of this task in Section~\ref{sec:preliminary}. Then, we introduce the proposed ATFM in Section~\ref{sec:acfm} and develop two unified frameworks to forecast short-term/long-term traffic flow in Section~\ref{sec:framework}. Extensive evaluation and comparisons are conducted in Section~\ref{sec:experiment}. Finally, we conclude this paper in Section~\ref{sec:conclusion}.

\section{Related Work}\label{sec:review}
\subsection{Traffic Flow Analysis}
As a crucial task in ITS, traffic flow analysis has been studied for decades~\cite{williams1998urban,castro2009online} due to its wide applications in urban traffic management and public safety monitoring. Traditional approaches~\cite{shekhar2007adaptive,li2012prediction,lippi2013short} usually used time series models (e.g., Vector Auto-Regression~\cite{johansen1991estimation}, ARIMA~\cite{box2015time} and their variants~\cite{williams1998urban}) for traffic flow prediction. However, most of these earlier methods modeled the evolution of traffic flow for each individual location separately and cannot well capture the complex spatial-temporal dependency.

Recently, deep learning based methods have been widely used in various traffic-related tasks \cite{duan2016efficient,chen2017visual,fouladgar2017scalable,ke2017short,wei2018intellilight,liu2019crowd}. Inspired by these works, many researchers have attempted to address traffic flow prediction with deep learning algorithms.
For instance, {\color{black}an artificial neural network termed ST-ANN~\cite{zhang2017deep} was proposed to forecast traffic flow by extracting both the spatial (values of 8 regions in the neighborhood) and temporal (8 previous time intervals) features. In \cite{zhang2016dnn}, a DNN-based model DeepST was proposed to capture various temporal properties (i.e. temporal closeness, period and trend). In \cite{zhang2017deep}, a deep ST-ResNet framework was developed with ResNet~\cite{he2016deep} to leverage the temporal closeness, period and trend information for citywide traffic flow prediction. }
Xu et al.~\cite{xu2018predcnn} designed a cascade multiplicative unit to model the dependencies between multiple frames and applied it to forecast the future traffic flow. Zhao et al.~\cite{zhao2017lstm} proposed a unified traffic forecast model based on long short-term memory network for short-term traffic flow forecast. Geng et al.~\cite{geng2019spatiotemporal} developed a multi-graph convolution network to encode the non-Euclidean pair-wise correlations among regions for spatiotemporal forecasting.
Currently, to overcome the scarcity of traffic flow data, Wang et al.~\cite{wang2018crowd} proposed to learn the target city model from the source city model with a region based cross-city deep transfer learning algorithm. Yao et al.~\cite{yao2019learning} incorporate the meta-learning paradigm into networks to tackle the problem of traffic flow prediction for the cities with only a short period of data collection.
{\color{black}However, the above-mentioned algorithms have two major disadvantages. First, some of them~\cite{zhang2016dnn,zhang2017deep,xu2018predcnn} simply employed convolution operations to extract temporal features and could not fully explore the temporal patterns. Second, all of them neglected the dynamic dependencies of spatial regions and failed to capture complex spatial patterns. In contrast, our ATFM incorporates two progress ConvLSTM units with a spatial attention map to effectively learn dynamic spatial-temporal features.}

\subsection{Temporal Sequences Modeling}
Recurrent neural network~(RNN) is a special class of artificial neural network for temporal sequences modeling. As an advanced variation, Long Short-Term Memory Networks~(LSTM) enables RNNs to store information over extended time intervals and exploit longer-term temporal dependencies.
Recently, LSTM has been widely applied to various sequential prediction tasks, such as natural language processing~\cite{luong2015effective} and speech recognition~\cite{graves2013speech}. Many works in computer vision community \cite{kalchbrenner2017video,lotter2017deep,wang2017predrnn} also combined CNN with LSTM to model the spatial-temporal information and achieved substantial progress in various tasks, such as video prediction. {\color{black}For instance, in \cite{kalchbrenner2017video}, a Video Pixel Network (VPN) learned the temporal relationships of previous frames in video with ConvLSTM to forecast the content of the next several frames. A predictive neural network (PredNet~\cite{lotter2017deep}) used multiple LSTM-based layers to predict future frames in a video sequence, with each layer making local predictions and only forwarding deviations from those predictions to subsequent layers. PredRNN~\cite{wang2017predrnn} utilized some stacked spatial-temporal LSTM layers to memorize both spatial and temporal variations of input frames. Without doubts, these models can be implemented and retained to forecast traffic flow, but they mainly focus on temporal modeling and are not aware of the dynamic spatial dependencies of traffic flow.}

Inspired by the success of the aforementioned works, many researchers~\cite{tian2015predicting,fu2016using,mackenzie2018evaluation} have attempted to address traffic flow prediction with recurrent neural networks.
However, existing works simply apply LSTM to extract feature and also cannot fully model the spatial-temporal evolution of traffic flow. {\color{black}Thus, a comprehensive module that can simultaneously learn the dynamic dependencies of both spatial view and temporal view is extremely desired for traffic flow prediction.}

\subsection{Attention Mechanism}
Visual attention~\cite{sharma2015action,lu2016knowing} is a fundamental aspect of the human visual system, which refers to the process by which humans focus the computational resources of their brain's visual system to specific regions of the visual field while perceiving the surrounding world. It has been recently embedded in deep convolution networks or recurrent neural networks to adaptively attend on mission-related regions while processing feedforward operations. {\color{black}For instance, in the task of visual question answering, Xu and Saenko~\cite{xu2016ask} chose some question-related regions dynamically with spatial attention to answer the questions about a given image. In crowd counting, Liu et al.~\cite{liu2018crowd} utilized an attention mechanism to select some local regions of the input image and then conducted local density map refinement. Tay et al.~\cite{tay2019aanet} integrated person attributes and attribute attention maps into a classification framework to solve the person re-identification problem.
Inspired by these work, our ATFM computes the attention weights of spatial regions at each time intervals and incorporates two ConvLSTM units to dynamically learn the spatial-temporal representations. Thanks to this simple yet effective attention mechanism, our method can favorably model the dynamic spatial-temporal dependencies of traffic flow.
}

\section{Preliminaries}\label{sec:preliminary}
In this section, we first introduce some basic elements of traffic flow and then elaborate the definition of the traffic flow prediction problem.

\textbf{Region Partition}:
There are many ways to divide a city into multiple regions in terms of different granularities and semantic meanings, such as road network ~\cite{deng2016latent} and zip code tabular~\cite{liu2019contextualized}. In this work, we follow the previous work~\cite{zhang2016dnn} to partition a city into $h \times w$ non-overlapping grid map based on the longitude and latitude. Each rectangular grid represents a different geographical region in the city. All partitioned regions of Beijing and New York City are shown in Fig.\ref{fig:citywide_crow_flow_visual}. With this simple partition strategy, the raw mobility data could be easily transformed into a matrix or tensor, which is the most common format of input data of the deep neural networks.

\textbf{Traffic Flow Map}:
In some practical applications, we can extract a mass of taxicabs/bikes trajectories from GPS signals or mobile phone signals. With these  trajectories, we measure the number of vehicles/bikes entering or leaving a given region at each time interval, which are called as inflow and outflow in our work. For convenience, we denote the traffic flow map at the ${t^{th}}$ time interval of ${d^{th}}$ day as a tensor ${\bm{M_d^t} \in R^{2\times h\times w}}$, in which the first channel is the inflow and the second channel denotes the outflow. Some examples of traffic flow maps are visualized in Fig.~\ref{fig:Sequential-Attention}.

\textbf{External Factors}:
As mentioned in \cite{zhang2017deep}, traffic flow can be affected by many complex external factors. For example, a sudden rain may seriously affect the traffic flow evolution and people would gather in some commercial areas for celebration on New Year's Eve. In this paper, we also consider the effect of meteorology information and holiday information, {\color{black}and their detail descriptions on different benchmarks can be found in Section~\ref{sec:setting}.}

{\textit{i) Meteorological preprocessing}}: Some meteorology factors (e.g., weather condition, temperature and wind speed) can be collected from a public website Wunderground\footnote{\url{https://www.wunderground.com/}}. Specifically, the weather condition is categorized into multiple categories (e.g., sunny and rainy) and it is digitized with One-Hot Encoding~\cite{harris2010digital}, while temperature and wind speed are scaled into the range [0, 1] with a min-max linear normalization.

{\textit{ii) Holiday preprocessing}}: Multiple categories of holiday (e.g., Chinese Spring Festival and Christmas) can be acquired from a calendar and encoded into a binary vector with One-Hot Encoding. We concatenate all data of external factors to a 1D tensor. The tensor of external factors at the ${t^{th}}$ time interval of ${d^{th}}$ day is represented as ${\bm{E_d^t}}$ in the following sections.

\textbf{Traffic Flow Prediction}:
Given the historical traffic flow maps and data of external factors until the ${t^{th}}$ time interval of ${d^{th}}$ day, we aim to predict the traffic flow map ${M_d^{t+1}}$, which is called as short-term prediction in our work. Moreover, we also extend our model to conduct long-term prediction, in which we forecast the traffic flow at the next several time intervals.

\section{Attentive Traffic Flow Machine}\label{sec:acfm}
\begin{figure}[t]
  \begin{center}
     \includegraphics[width=0.975\columnwidth]{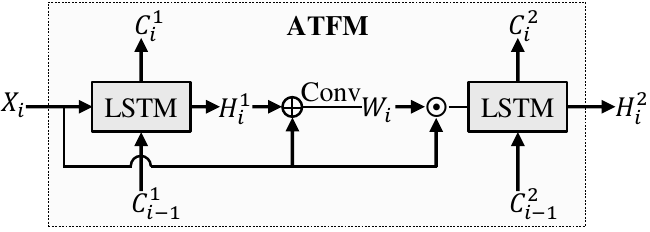}
  \vspace{-5mm}
  \end{center}
   \caption{Overview of the proposed Attentive Traffic Flow Machine (ATFM). ${X_i}$ is the normal traffic flow feature of the ${i^{th}}$ iteration. ``$\bigoplus$'' denotes a feature concatenation operation and ``$\odot$'' refers to an element-wise multiplication operation. The first ConvLSTM unit takes ${X_i}$ as input and incorporates the historical information to dynamically generate a spatial attention map ${W_i}$. The second ConvLSTM unit learns a more effective spatial-temporal feature representation from the attentionally weighted traffic flow features.}
\vspace{0mm}
\label{fig:acfm}
\end{figure}

In this section, we propose a unified neural network, named Attentive Traffic Flow Machine (ATFM), to learn the spatial-temporal representations of traffic flow. ATFM is designed to adequately capture various contextual dependencies of the traffic flow, e.g., the spatial consistency and the temporal dependency of long and short term. As shown in Fig.~\ref{fig:acfm}, the proposed ATFM consists of two progressive ConvLSTM units connected with a convolutional layer for attention weight prediction at each time step. Specifically, the first ConvLSTM unit learns temporal dependency from the normal traffic flow features, the extraction process of which is described in Section~\ref{sec:feature}. The output hidden state encodes the historical evolution information and it is concatenated with the current traffic flow feature for spatial weight map inference. The second ConvLSTM unit takes the re-weighted traffic flow features as input at each time-step and is trained to recurrently learn the spatial-temporal representations for further traffic flow prediction.

Let us denote the input feature of traffic map at the ${i}^{th}$ iteration as ${X_i \in R^{c\times h\times w}}$, with $h$, $w$ and $c$ representing the height, width and the number of channels. At each iteration, the first ConvLSTM unit takes ${X_i}$ as input and updates its memorized cell state $C^1_i$ with an input gate $\mathcal{I}^1_{i}$ and a forget gate $\mathcal{F}^1_{i}$. Meanwhile, it updates its new hidden state $H^1_{i}$ with an output gate $\mathcal{O}^1_{i}$. The computation process of our first ConvLSTM unit is formulated as:
{
\begin{equation}
\begin{split}
\mathcal{I}^1_{i}=&\sigma \left ( w_{xi}\ast {X}_{i}+w_{hi}\ast H^1_{i-1}+w_{ci} \boldsymbol{\odot} C^1_{i-1}+ b_{i} \right )\\
\mathcal{F}^1_{i}=&\sigma \left ( w_{xf}\ast {X}_{i}+w_{hf}\ast H^1_{i-1}+w_{cf} \boldsymbol{\odot} C^1_{i-1}+ b_{f} \right )\\
C^1_{i}=&\mathcal{F}^1_{i} \boldsymbol{\odot} C^1_{i-1}+\mathcal{I}^1_{i} \boldsymbol{\odot} tanh \left ( w_{xc}\ast {X}_{i}+w_{hc}\ast H^1_{i-1}+ b_{c} \right )\\
\mathcal{O}^1_{i}=&\sigma \left ( w_{xo}\ast {X}_{i}+w_{ho}\ast H^1_{i-1}+w_{co} \boldsymbol{\odot} C^1_{i}+ b_{0} \right )\\
H^1_{i}=&\mathcal{O}^1_{i} \boldsymbol{\odot} tanh\left ( C^1_{i} \right )
\end{split}
\label{lstm}
\end{equation}
}%
where $w_{\alpha \beta}\left (\alpha \in\left \{ x,h,c\right \} ,\beta\in\left \{ i,f,o,c\right \}\right )$ are the parameters of convolutional layers in ConvLSTM. ${\sigma}$ denotes the logistic sigmoid function and $\boldsymbol{\odot}$ is an element-wise multiplication operation.
For notation simplification, we denote Eq.(\ref{lstm}) as:
\begin{equation}
\label{equ:first_lstm}
H_i^1, C_i^1 = \textbf{ConvLSTM}(H_{i-1}^1, C_{i-1}^1, X_i).
\end{equation}%
Generated from the memorized cell state $C^1_i$, the new hidden state ${H_i^1}$ encodes the dynamic evolution of historical traffic flow in temporal view.

We then integrate a deep attention mechanism to dynamically model the spatial dependencies of traffic flow. Specifically, we incorporate the historical state ${H_i^1}$ and current state ${X_i}$ to infer an attention map ${W_i}$, which is implemented by:
\begin{equation}
\label{equ:attention}
W_i = \textbf{Conv}_{1\times 1}(H_i^1 \oplus X_i, w_a),
\end{equation}%
where $\oplus$ denotes a feature concatenation operation and $w_a$ is the parameters of a convolutional layer with a kernel size of ${1\times 1}$. The attention map ${W_i}$ is learned to discover the weights of each spatial location on the input feature map ${X_i}$.

Finally, we learn a more effective spatial-temporal representation with the guidance of attention map.
After reweighing the normal traffic flow feature map by multiplying ${X_i}$ and ${W_i}$ element by element, we feed it into the second ConvLSTM unit and generate a new hidden state ${H_i^2 \in R^{c\times h\times w}}$, which is expressed as:
\begin{equation}
\label{equ:second_lstm}
H_i^2, C_i^2 = \textbf{ConvLSTM}(H_{i-1}^2, C_{i-1}^2, X_i \boldsymbol{\odot} W_i),
\end{equation}%
where ${H_i^2}$ encodes the attention-aware content of current input as well as memorizes the contextual knowledge of previous moments. When the elements in a sequence of traffic flow maps are recurrently fed into ATFM, the last hidden state encodes the information of the whole sequence and it can be used as the spatial-temporal representation for evolution analysis of future flow map.

\begin{figure*}[t]
  \begin{center}
     \includegraphics[width=1.8\columnwidth]{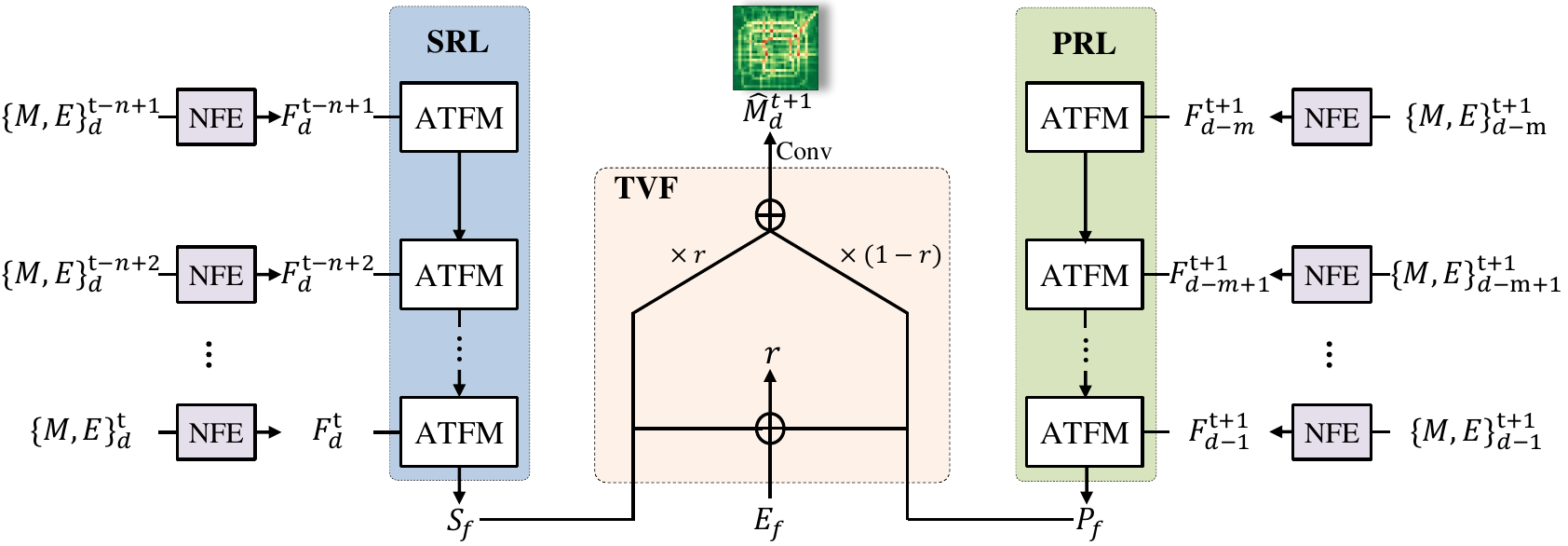}
  \vspace{-5mm}
  \end{center}
   \caption{The architecture of SPN based on ATFM for citywide short-term traffic flow prediction. It consists of four components: (1) a normal feature extraction ({\bf{NFE}}) module, (2) a sequential representation learning ({\bf{SRL}}) module, (3) a periodic representation learning ({\bf{PRL}}) module and (4) a temporally-varying fusion ({\bf{TVF}}) module. $\{M,E\}^i_j$ denotes the  traffic flow map ${M^i_j}$ and external factors tensor ${E^i_j}$ at the ${i^{th}}$ time interval of the ${j^{th}}$ day. ${F^i_j}$ is the embedded feature of ${M^i_j}$ and ${E^i_j}$. ${S_{f}}$ and ${P_{f}}$ are sequential representation and periodic representation, while external factors integrative feature ${E_{f}}$ is the element-wise addition of external factors features of all relative time intervals. ``$\bigoplus$'' refers to feature concatenation. The symbols ${r}$ and ${(1-r)}$ reflect the importance of ${S_{f}}$ and ${P_{f}}$ respectively. ${\widehat{M}^{t+1}_d}$ is the predicted traffic flow map.}
\vspace{0mm}
\label{fig:network-structure}
\end{figure*}

\section{Citywide Traffic Flow Prediction}\label{sec:framework}
In this section, we first develop a deep neural network framework which incorporates the proposed ATFM for citywide short-term traffic flow prediction. We then extend this framework to predict long-term traffic flow with an extra LSTM prediction network. {\color{black}Notice that our framework is general and can be applied for other traffic prediction tasks, such as the citywide passenger demand prediction described in Section~\ref{sec:demand}.}

\subsection{Short-term Prediction}
As illustrated in Fig.~\ref{fig:network-structure}, our short-term prediction framework consists of four components: (1) a normal feature extraction ({\bf{NFE}}) module, (2) a sequential representation learning ({\bf{SRL}}) module, (3) a periodic representation learning ({\bf{PRL}}) module and (4) a temporally-varying fusion ({\bf{TVF}}) module.
First, the NFE module is used to extract the normal features of traffic flow map and external factors tensor at each time interval.
Second, the SRL and PRL modules are employed to model the contextual dependencies of traffic flow at two different temporal scales.
Third, the TVF module adaptively merges the feature representations of SRL and PRL with the fused weight learned from the comprehensive features of various factors.
Finally, the fused feature map is fed to one additional convolution layer for traffic flow map inference.
For convenience, this framework is denoted as \textbf{S}equential-\textbf{P}eriodic \textbf{N}etwork (\textbf{SPN}) in following sections.

\begin{figure}[ht]
  \begin{center}
     \includegraphics[width=1\columnwidth]{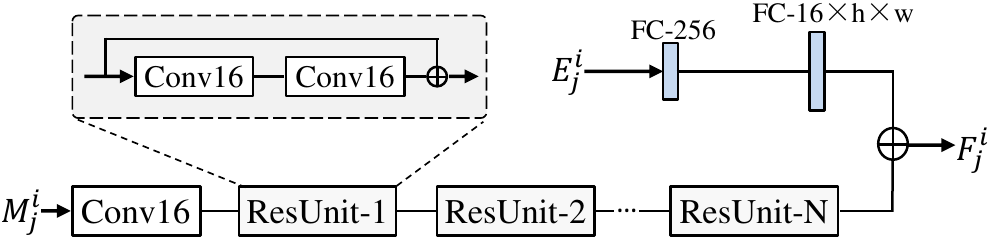}
  \vspace{-9mm}
  \end{center}
   \caption{The architecture of the subnetwork for normal feature extraction. It is designed as a concatenation of the embedded traffic flow feature and the external factor feature. 
   Conv16 is a convolutional layer with 16 channels and FC-$k$ denotes a fully-connected layer with $k$ output neurons.}
\vspace{0mm}
\label{fig:ResNet}
\end{figure}

\subsubsection{\bf{Normal Feature Extraction}}\label{sec:feature}
We first describe how to extract the normal features of traffic flow and external factors, which will be further fed into the SRL and PRL modules for dynamic spatial-temporal representation learning.

As shown in Fig.~\ref{fig:ResNet}, we utilize a customized ResNet~\cite{he2016deep} to learn feature embedding from the given traffic flow map ${M^i_j}$. Specifically, our ResNet consists of ${N}$ residual units, each of which has two convolutional layers with channel number of 16 and kernel size of ${3\times3}$. To maintain the resolution $h\times w$, we set the stride of all convolutional layers to 1 and do not adopt any pooling layers in ResNet. Following \cite{zhang2017deep}, we first scale ${M^i_j}$ into the range $[-1,1]$ with a min-max linear normalization and then feed it into the ResNet to generate the traffic flow feature, which is denoted as ${F^i_j(M) \in R^{16\times h\times w}}$.

Then, we extract the feature of the given external factors tensor ${E^i_j}$ with a Multilayer Perceptron. We implement it with two fully-connected layers. The first layer has 40 output neurons and the second one has ${16 \times  h \times  w}$ output neurons. We reshape the output of the last layer to form the 3D external factor feature ${F^i_j(E) \in R^{16\times h\times w}}$.
Finally, we fuse ${F^i_j(M)}$ and ${F^i_j(E)}$ to generate an embedded feature ${F_j^i}$, which is expressed as:
\begin{equation}
F_j^i= F^i_j(M) \oplus F^i_j(E),
\end{equation}%
where $\oplus$ denotes feature concatenation. $F_j^i$ is the normal feature at a specific time interval and it is unaware of the dynamic spatial dependencies of traffic flow. Thus, the following two modules are proposed to dynamically learn the spatial-temporal representation.

\begin{figure*}[t]
  \begin{center}
     \includegraphics[width=1.80\columnwidth]{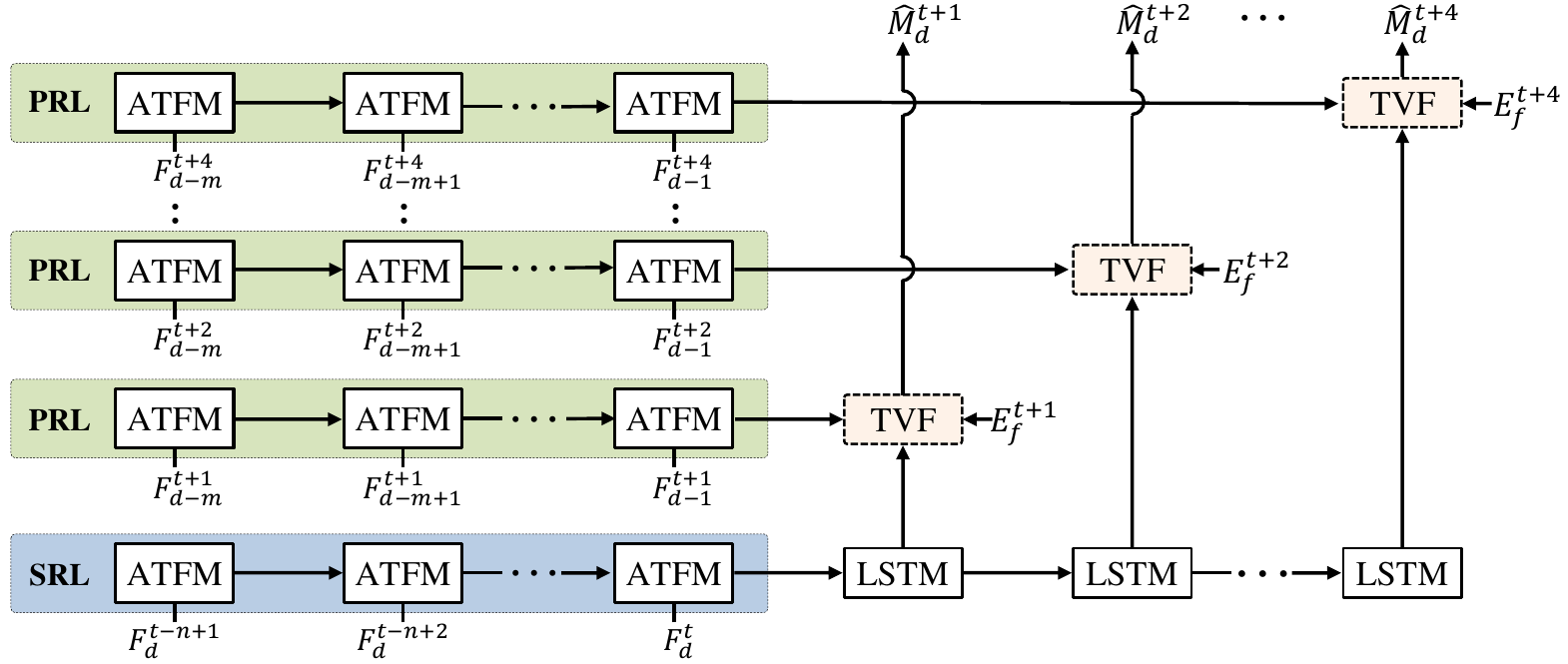}
  \vspace{-6mm}
  \end{center}
   \caption{The architecture of the SPN-LONG for Long-term Traffic Flow Prediction. $F^i_j$ is the normal traffic flow feature described in Section~\ref{sec:feature}. $E_f^{t+i}$ is the element-wise addition of external factors features ${\{E^{t-k}_{d} \big| k = n-1, ... ,0\}}$ and ${\{E^{t+i}_{d-k} \big| k = m,...,1\}}$. }
\vspace{0mm}
\label{fig:SPN-Long-structure}
\end{figure*}

\subsubsection{\bf{Sequential Representation Learning}}\label{sec:srl}
The evolution of citywide traffic flow is usually affected by the recent traffic states. For instance, a traffic accident occurring on a main road of the studied city during morning rush hours may seriously affect the traffic flow of nearby regions in subsequent time intervals. In this subsection, we develop a sequential representation learning (SRL) module based on the proposed ATFM to fully model the evolution trend of traffic flow.

First, we take the normal traffic flow features of several recent time intervals to form a group of sequential temporal features, which is denoted as:
\begin{equation}
S_{in} = \{F^{t-k}_d \big| k = n-1, n-2,...,0\},
\end{equation}
where ${n}$ is the length of the sequentially related time intervals.
We then apply the proposed ATFM to learn sequential representation from the temporal features $S_{in}$. As shown on the left of Fig.~\ref{fig:network-structure}, at each iteration, ATFM takes one element in $S_{in}$ as input and learns to selectively memorize the spatial-temporal context of the sequential traffic flow.
Finally, we get the sequential representation ${S_{f} \in R^{16\times h\times w}}$ by feeding the last hidden state of ATFM into a $1\times1$ convolution layer. ${S_{f}}$ encodes the sequential evolution trend of traffic flow.

\subsubsection{\bf{Periodic Representation Learning}}
In urban transportation systems, there exist some periodicities which make a significant impact on the changes of traffic flow. For example, the traffic conditions are very similar during morning rush hours of consecutive workdays, repeating every 24 hours. Thus, in this subsection, we propose a periodic representation learning (PRL) module that fully captures the periodic dependencies of traffic flow with the proposed ATFM.

Similar to the sequential representation learning, we first construct a group of periodic temporal features
\begin{equation}
P_{in} = \{F^{t+1}_{d-k} \big| k = m, m-1,...,1\},
\end{equation}
where ${n}$ is the length of the periodic days.
At each iteration, we feed one element in $P_{in}$ into ATFM to dynamically learn the periodic dependencies, as shown on the right of Fig.~\ref{fig:network-structure}. After the last iteration, we feed the hidden state of ATFM into a $1\times1$ convolutional layer to generate the final periodic representation ${P_{f} \in R^{16\times h\times w}}$. Encoding the periodic evolution trend of traffic flow, ${P_{f}}$ is proved to be effective for traffic prediction in our experiments.

\subsubsection{\bf{Temporally-Varying Fusion}}\label{sec:fusion}
As described in the two previous modules, the future traffic flow is affected by the sequential representation ${S_{f}}$ and the periodic representation ${P_{f}}$ simultaneously. We find that the relative importance of these two representations is temporally dynamic and it is suboptimal to directly concatenate them without any specific preprocessing.
To address this issue, we propose a novel temporally-varying fusion (TVF) module to adaptively fuse the representations ${S_{f}}$ and ${P_{f}}$ with different weights learned from the comprehensive features of various internal and external factors.

In TVF module, we take the sequential representation ${S_{f}}$, the periodic representation ${P_{f}}$ and the external factors integrative feature ${E_{f}}$ to determine the fusion weight. Specifically, ${E_{f}}$ is the element-wise addition of the external factors features $\{F(E)^{t-k}_d \big| k = n-1, n-2,...,0\}$ and $\{F(E)^{t+1}_{d-k} \big| k = m, m-1,...,1\}$.
As shown in Fig.~\ref{fig:network-structure}, we first feed the concatenation of ${S_{f}}$, ${P_{f}}$ and ${E_{f}}$ into two fully-connected layers for fusion weight inference. The first layer has 32 output neurons and the second one has only one neuron. We then obtain the fusion weight of ${S_{f}}$ by applying a sigmoid function on the output of the second FC layer. The weight of ${P_{f}}$ is automatically set to $1-r$. We then fuse these two temporal representations on the basis of the learned weights and compute a comprehensive spatial-temporal representation $\text{SP}_f$ as:
\begin{equation}
\label{equ:fusion}
\text{SP}_f= r\times S_{f} \oplus (1-r)\times P_{f},
\end{equation}
where $\text{SP}_f$ contains the sequential and periodic dependencies of traffic flow.

Finally, we feed $\text{SP}_f$ into a $1\times1$ convolutional layer with two filters to predict the future traffic flow map ${\widehat{M}^t_d \in R^{2\times h\times w}}$ with following formula:
\begin{equation}
\widehat{M}^t_d = tanh(\text{SP}_f * w_p).
\end{equation} %
where $w_p$ is the parameters of the predictive convolutional layer and the hyperbolic tangent ${tanh}$ ensures the output values are within the range $[-1, 1]$. Further, the predicted map $\widehat{M}^t_d$ is re-scaled back to normal value with an inverted min-max linear normalization.

\begin{table*}[t]
  \caption{The Overview of TaxiBJ and BikeNYC datasets. ``\# Taxis/Bikes" denotes the number of taxis or bikes in the datasets. Other texts with ``\#'' have similar meanings.}
  \vspace{-1mm}
  \centering
    \begin{tabular}{c|c|c|c}
    \hline
    \multicolumn{2}{c|}{\textbf{Dataset}}  & \textbf{TaxiBJ} & \textbf{BikeNYC} \\
    \hline\hline
    \multirow{10}{*}{\textbf{Traffic Flow}}  & City & Beijing & New York \\
    \cline{2-4}
      & Gird Map Size & (32, 32) & (16, 8) \\
    \cline{2-4}
      & Data Type & Taxi GPS & Bike Rent\\
    \cline{2-4}
      & \multirow{4}*{Time Span} & {~}7/1/2013 - 10/30/2013 & \multirow{4}*{4/1/2014 - 9/30/2014 }\\
      & & {~}3/1/2014 - {~}6/30/2014 & \\
      & & {~}3/1/2015 - {~}6/30/2015&\\
      & & 11/1/2015 - {~}4/10/2016&\\
    \cline{2-4}
      & \# Taxis/Bikes & 34,000+ & 6,800+ \\
    \cline{2-4}
      & Time Interval & 0.5 hour & 1 hour \\
    \cline{2-4}
      & \# Available Time Interval & 22,459  & 4,392 \\
    \hline

    \multirow{4}{*}{\textbf{External Factors}}   &  \# Holidays & 41 & 20  \\
    \cline{2-4}
      & \multirow{2}{*}{Weather Conditions} & 16 types & \multirow{2}{*}{$\setminus$} \\
      & & (e.g., Sunny, Rainy) & \\
    \cline{2-4}
      & Temperature / $^\circ$C  & $[-24.6, 41.0]$ & $\setminus$ \\ 
    \cline{2-4}
      & Wind Speed / mph & $[0, 48.6]$ & $\setminus$ \\
    \hline
    \end{tabular}
  \label{tab:dataset_detail}
\end{table*}

\subsection{Long-term Prediction}
In this subsection, we extend our method to predict the longer-term traffic flow. With a similar setting of short-term prediction, we incorporate the sequential data and periodic data at previous time intervals to forecast the traffic flow at the next four time intervals. For convenience, we denote this model as SPN-LONG in the following sections.

The architecture of our SPN-LONG is shown in Fig~\ref{fig:SPN-Long-structure}.
For each previous time interval, we first extract its normal features $F^i_j$ with the proposed NFE module.
Then, the features in $\{F^{t-k}_d \big| k = n-1, n-2,...,0\}$ are recurrently fed into ATFM to learn the sequential representation.
The output sequential representation is then fed into a LSTM prediction network. With four ConvLSTM units, this prediction network is designed to forecast the traffic flow at the next four time intervals. Specifically, at $i^{th}$ LSTM, we use a TVF module to adaptively fuse its hidden state and the periodic representation learned from $\{F^{t+i}_{d-k} \big| k = m,...,1\}$. The external factors integrative feature $E_f^{t+i}$ is the element-wise addition of ${\{E^{t-k}_{d} \big| k = n-1,...,0\}}$ and ${\{E^{t+i}_{d-k} \big| k = m,...,1\}}$. Finally, we take the output of $i^{th}$ TVF module to predict $\widehat{M}^{t+i}_d$ with a convolutional layer.

\section{Experiments}\label{sec:experiment}
In this section, we first introduce the commonly-used benchmarks and evaluation metrics of citywide traffic flow prediction. Then, we compare the proposed approach with several state-of-the-art methods under different settings. Furthermore, we conduct extensive component analysis to demonstrate the effectiveness of each part in our model. Finally, we apply the proposed method to passenger pickup/dropoff demands forecasting and show its generalization for other traffic prediction tasks.

\subsection{Experimental Setting}\label{sec:setting}

\subsubsection{\textbf{Dataset}}
In this work, we forecast the inflow and outflow of citywide transportation entities on two representative benchmarks, including the TaxiBJ dataset~\cite{zhang2017deep} for taxicab flow prediction and the BikeNYC dataset~\cite{zhang2016dnn} for bike flow prediction. {\color{black}These two datasets can be accessed publicly and various comparison algorithms can be evaluated on the same testing sets for fair comparisons.} The summaries of TaxiBJ and BikeNYC are shown in Table~\ref{tab:dataset_detail}\footnote{The details of TaxiBJ and BikeNYC dataset are from quoted from \cite{zhang2017deep}.}.

{\textit{TaxiBJ Dataset:}} In this dataset, a mass of taxi GPS trajectories are collected from 34 thousand taxicabs in Beijing for over 16 months. The time interval is half an hour and 22,459 traffic flow maps with size ${32\times32}$ are generated from these trajectory data. The external factors contain weather conditions, temperature, wind speed and 41 categories of holidays. This dataset is officially divided into a training set and a testing set. The number of testing data is around 6\% of that of training data. Specifically, the data in the last four weeks are used for evaluation and the rest data are used for training.

{\textit{BikeNYC Dataset:}} Generated from the NYC bike trajectory data for 182 days, this dataset contains 4,392 traffic flow maps with a time interval of one hour and the size of these maps is ${16\times8}$. As for external factors, 20 holiday categories are recorded. This dataset has the similar training-testing ratio of TaxiBJ. Specifically, the data of the first 172 days are used for training and the data of the last ten days are chosen to be the testing set.

\subsubsection{\textbf{Implementation Details}}
We adopt the PyTorch~\cite{paszke2017automatic} toolbox to implement our traffic flow prediction network. The sequential length $n$ and the periodic length $m$ are set to 4 and 2, respectively. For fair comparison with ST-ResNet~\cite{zhang2017deep}, we develop the customized ResNet in Section~\ref{sec:feature} with 12 residual units on the TaxiBJ dataset and 4 residual units on the BikeNYC dataset.
The filter weights of all convolutional layers and fully-connected layers are initialized by Xavier~\cite{glorot2010understanding}.
The size of a minibatch is set to 64 and the learning rate is ${10^{-4}}$. We optimize the parameters of our network in an end-to-end manner via Adam optimization~\cite{kingma2014adam} by minimizing the Euclidean loss.

\subsubsection{\textbf{Evaluation Metric}}
In traffic flow prediction, Root Mean Square Error (RMSE) and Mean Absolute Error (MAE) are two popular evaluation metrics used to measure the performances of related methods. Specifically, they are defined as:
\begin{equation}
\text{RMSE} = \sqrt{{\frac{1}{z}}{\sum_{i=1}^z{(\widehat{Y_i} - Y_i)}^2}},
{~~~}
\text{MAE} = {\frac{1}{z}}{\sum_{i=1}^z |\widehat{Y_i} - Y_i|}
\end{equation}
%
where ${\widehat{Y_i}}$ and ${Y_i}$ represent the predicted flow map and its ground truth, respectively. $z$ indicates the number of samples used for validation. Noted that some partitioned regions in New York City are water areas and their flow are always zero, which may decrease the mean error and affect the evaluation of algorithm performance. To correctly reflect the performance of different methods on BikeNYC dataset, we re-scale their mean errors with a ratio (1.58) provided by ST-ResNet.

\subsection{Comparison for Short-term Prediction}
In this subsection, we compare the proposed method with ten typical methods for short-term traffic flow prediction. These compared methods can be divided into three categories, including: {\bf{(i)}} traditional models for time series forecasting, {\bf{(ii)}} deep learning networks particularly designed for traffic flow prediction and {\bf{(iii)}} the state-of-the-art approaches originally designed for some related tasks. The details of the compared methods are described as follows.
\begin{itemize}
\item \textbf{HA:} Historical Average (HA) is a simple model that directly predicts the future traffic flow by averaging the historical flow in the corresponding periods. For example, the predicted flow at 7:00 am to 7:30 am on a specific Tuesday is the average flow from 7:00 am to 7: 30 am on all historical Tuesdays.
\item \textbf{ARIMA~\cite{box2015time}:} Auto-Regressive Integrated Moving Average (ARIMA) is a famous statistical analysis model that uses time series data to predict future trends.
\item \textbf{SARIMA~\cite{williams1998urban}:} Seasonal ARIMA (SARIMA) is an advanced variant of ARIMA that considers the seasonal terms.
\item \textbf{VAR~\cite{johansen1991estimation}:} Vector Auto-Regression (VAR) is a well-known stochastic process model and it can capture the linear interdependencies among multiple time series.
\item \textbf{DeepST~\cite{zhang2016dnn}:} This is a DNN-based model and it utilizes various temporal properties to conduct prediction.
\item \textbf{ST-ANN~\cite{zhang2017deep}:} As an artificial neural network, this model extracts spatial (8 nearby region values) and temporal (8 previous time intervals) features for future traffic flow prediction.
\item \textbf{ST-ResNet~\cite{zhang2017deep}:} As an advanced version of DeepST, this model incorporates the closeness, period, trend data as well as external factors to predict traffic flow with residual networks.
\item \textbf{VPN~\cite{kalchbrenner2017video}:} Video Pixel Networks (VPN) is a probabilistic video model designed for multi-frames prediction. A variant of VPN based on RMBs is implemented for traffic flow prediction.
\item \textbf{PredNet~\cite{lotter2017deep}:} As a predictive neural network, this model is originally developed to predict the content of subsequent frame in a video sequence. We apply this method to traffic flow prediction.
\item \textbf{PredRNN~\cite{wang2017predrnn}:} This method is also originally designed for video generation and it is implemented by memorizing both spatial and temporal variations of input frames with a predictive recurrent neural network for future frames generation. In this work, it is re-implemented to forecast traffic flow.
\end{itemize}

\begin{table}[t]
 \caption{Quantitative comparisons on TaxiBJ and BikeNYC. Our method outperforms the existing methods on both datasets.}
  \vspace{0mm}
\newcommand{\tabincell}[2]{\begin{tabular}{@{}#1@{}}#2\end{tabular}}
  \centering
    \begin{tabular}{c|c|c|c|c}
    \hline
    \multirow{2}{*}{{Method}} & \multicolumn{2}{c|}{\textbf{TaxiBJ}} & \multicolumn{2}{c}{\textbf{BikeNYC}} \\
    \cline{2-5}
    & RMSE & MAE & RMSE & MAE \\
    \hline
    HA & 57.79 & - & 21.57 & - \\
    \hline
    SARIMA & 26.88 & - & 10.56 & - \\
    \hline
    VAR    & 22.88 & - & 9.92 & - \\
    \hline
    ARIMA  & 22.78 & - & 10.07 & - \\
    \hline
    ST-ANN & 19.57 & - & - &- \\
    \hline
    DeepST & 18.18 & - & 7.43 & - \\
    \hline
    VPN & 16.75 & 9.62 & 6.17 & 3.68 \\
    \hline
    ST-ResNet & 16.69 & 9.52 & 6.37 & 2.95 \\
    \hline
    PredNet & 16.68 & 9.67 & 7.45 & 3.71 \\
    \hline
    PredRNN & 16.34 & 9.62 & 5.99 & 4.89 \\
    \hline
    SPN (Ours) & \textbf{\color{red}15.31} & \textbf{\color{red}9.14} & \textbf{\color{red}5.59} & \textbf{\color{red}2.74} \\
    \hline
    \end{tabular}
  \label{tab:BJ_NYC_result}
\end{table}

{\bf{Comparison on All Time Intervals:}} The performance of the proposed method and the other ten compared methods are summarized in Table~\ref{tab:BJ_NYC_result}.
{\color{black}Among these methods, the baseline model is HA that obtains a RMSE of 57.79 on the TaxiBJ dataset and 21.57 on the BikeNYC dataset. Although having some progress, the traditional time series algorithms (e.g., VAR, ARIMA, and SARIMA) still show inferior performance on both datasets, since these shallow models rely on handcrafted features and have weak capacity to model complex patterns. Thanks to the deep representation learning, the recent CNN-based methods ST-ANN, DeepST, and ST-ResNet can decrease the errors to some extent. For instance, ST-ResNet reduces the RMSE to 16.59 on TaxiBJ and to 6.37 on BikeNYC. However, only with CNN features, these models fail to fully capture the temporal patterns. When applying recurrent neural networks to model the temporal evolution of traffic flow, the RNN-based methods VPN, PredNet and PredRNN can defeat the aforementioned CNN-based models. Nevertheless, the dynamic spatial dependencies of traffic flow are neglected in these methods and this task still cannot be solved perfectly.}
In contrast, our method can further improve the performance by explicitly learning the spatial-temporal feature and dynamically modeling the spatial attention weighting of each spatial influence. Specifically, our method achieves a RMSE of 15.31 on the TaxiBJ dataset, outperforming the previous best approach PredRNN by 6.3\% relatively. {\color{black}On the BikeNYC dataset, our method also boosts the highest prediction accuracy},~i.e.,~decreases the RMSE from 5.99 to 5.59, and outperforms other methods.

{\color{black}Notice that the official BikeNYC dataset does not contain meteorological information. To enrich the external factors of BikeNYC, we collect the information of weather conditions (31 types), temperature ($[-1.1, 33.9]$) and wind speed ($[0, 33]$) from the popular meteorological website Wunderground. That meteorological information is processed with the same technique described in Section~\ref{sec:preliminary}. After combining these factors, our method further decreases the RMSE and MAE to 5.50 and 2.71, respectively. For fair comparison with other methods, we mainly report the performance trained with the official BikeNYC dataset in the following sections.
}

\begin{figure}[t]
  \begin{center}
     \includegraphics[width=1\columnwidth]{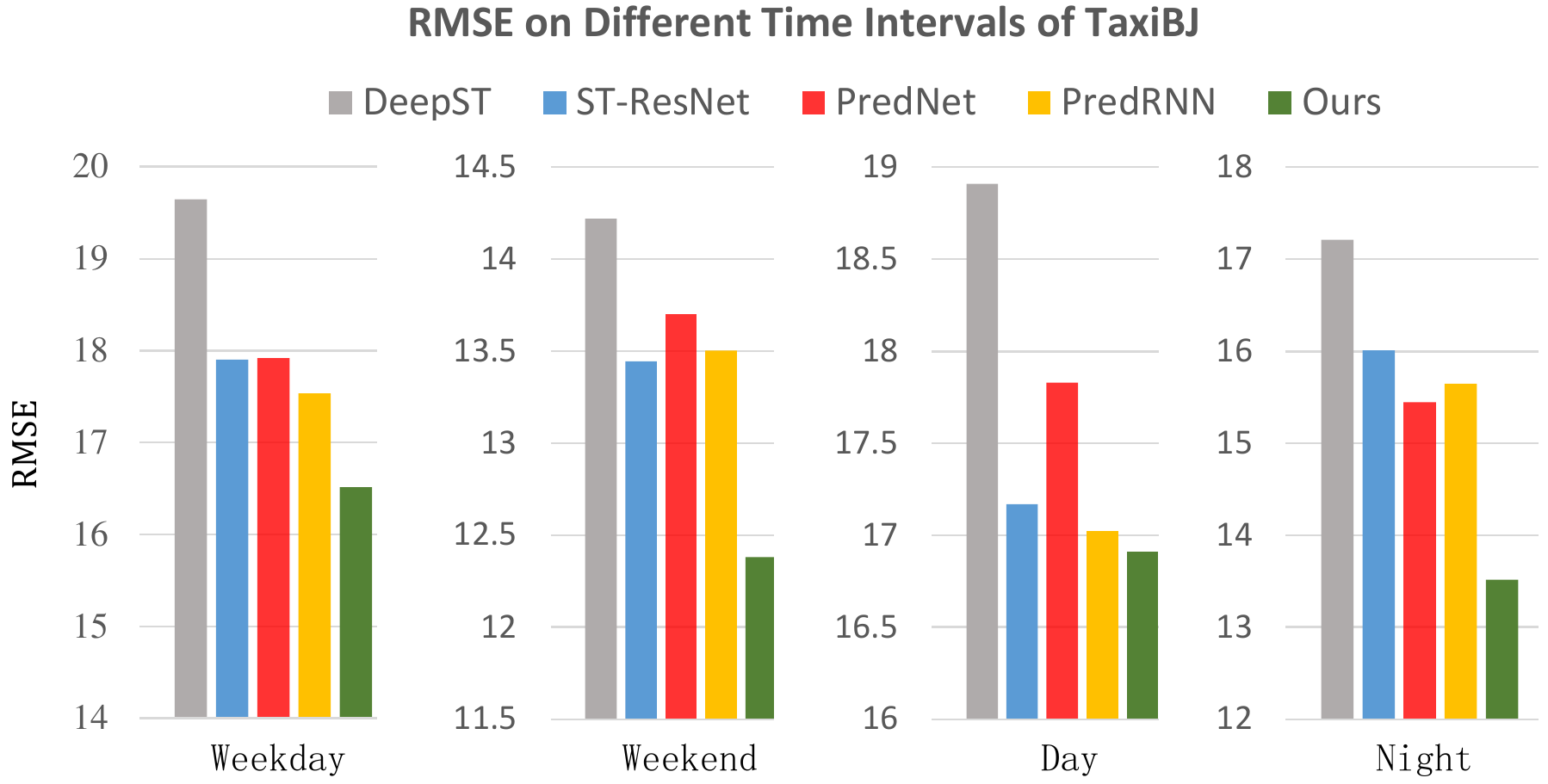}
  \vspace{-9mm}
  \end{center}
   \caption{The RMSE of weekday, weekend, day and night {\color{black}on the TaxiBJ dataset}. The weekday RMSE is the average result from Monday to Friday, while the weekend RMSE is the average result of Saturday and Sunday. The day RMSE and the night RMSE are the average result from 6:00 to 18:00 and from 18:00 to 6:00, respectively. Best view in color.}
\vspace{0mm}
\label{fig:day_night}
\end{figure}

\begin{figure}[t]
  \begin{center}
     \includegraphics[width=1\columnwidth]{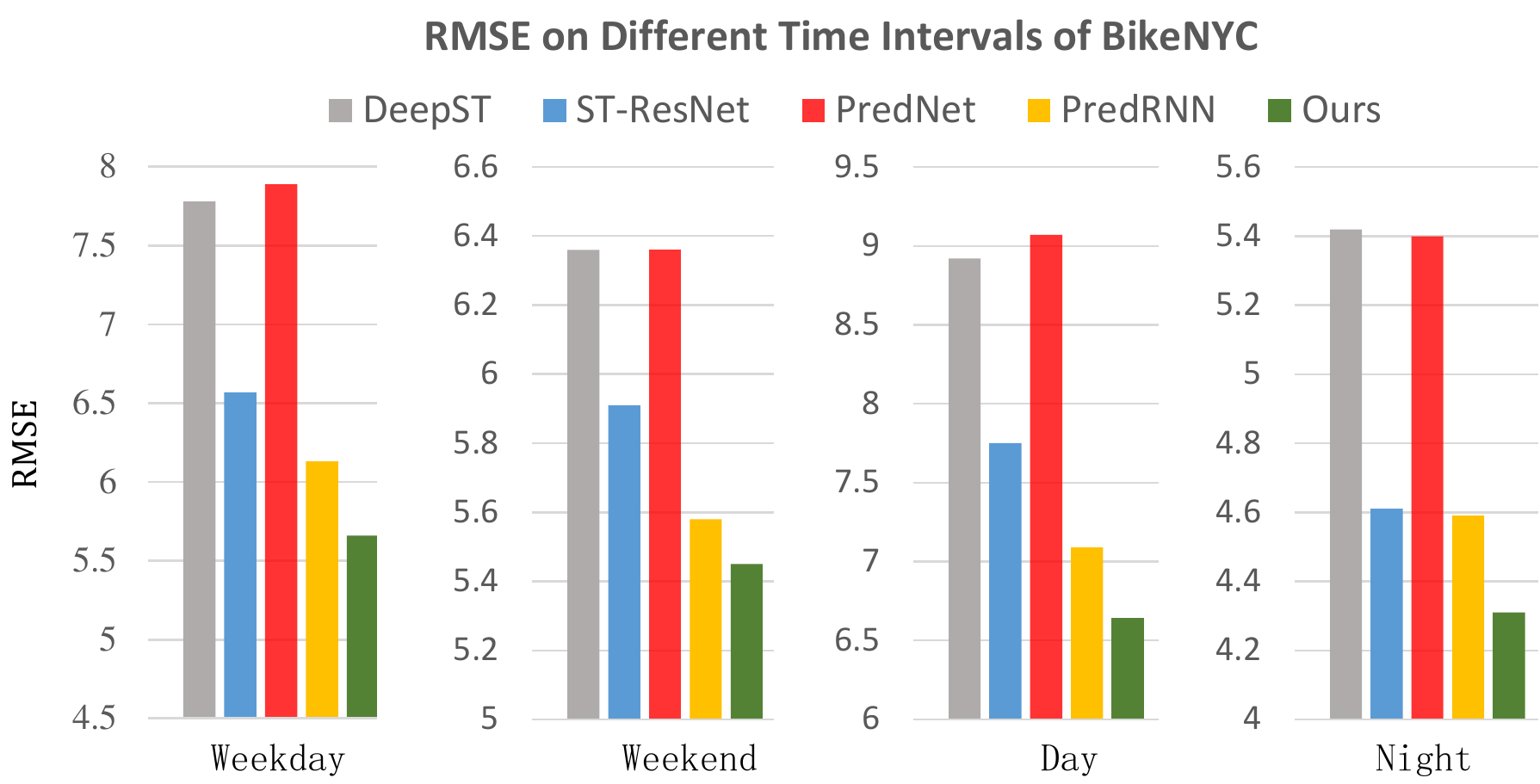}
  \vspace{-9mm}
  \end{center}
   \caption{{\color{black}The RMSE of weekday, weekend, day and night on the BikeNYC dataset}.}
\vspace{0mm}
\label{fig:day_night_NYC}
\end{figure}

{\bf{Comparison on Different Time Intervals:}} {\color{black}As previously described, traffic flow is time-varying and its temporal patterns are very complex. To explore the model's stability}, we compare the performance of five deep learning-based methods at different time intervals, such as weekday (from Monday to Friday), weekend (Saturday and Sunday), day (from 6:00 to 18:00) and night (from 18:00 to 6:00). As shown in Fig.~\ref{fig:day_night} and Fig.~\ref{fig:day_night_NYC}, our method outperforms other compared methods under various settings {\color{black}on both TaxiBJ and BikeNYC, since our ATFM can effectively learn the temporal patterns of traffic flow and the Temporally-Varying Fusion module can flexibly combine the information of different temporal sequences.} These experiments well demonstrate the robustness of our method.

\begin{figure}[t]
  \begin{center}
     \includegraphics[width=1\columnwidth]{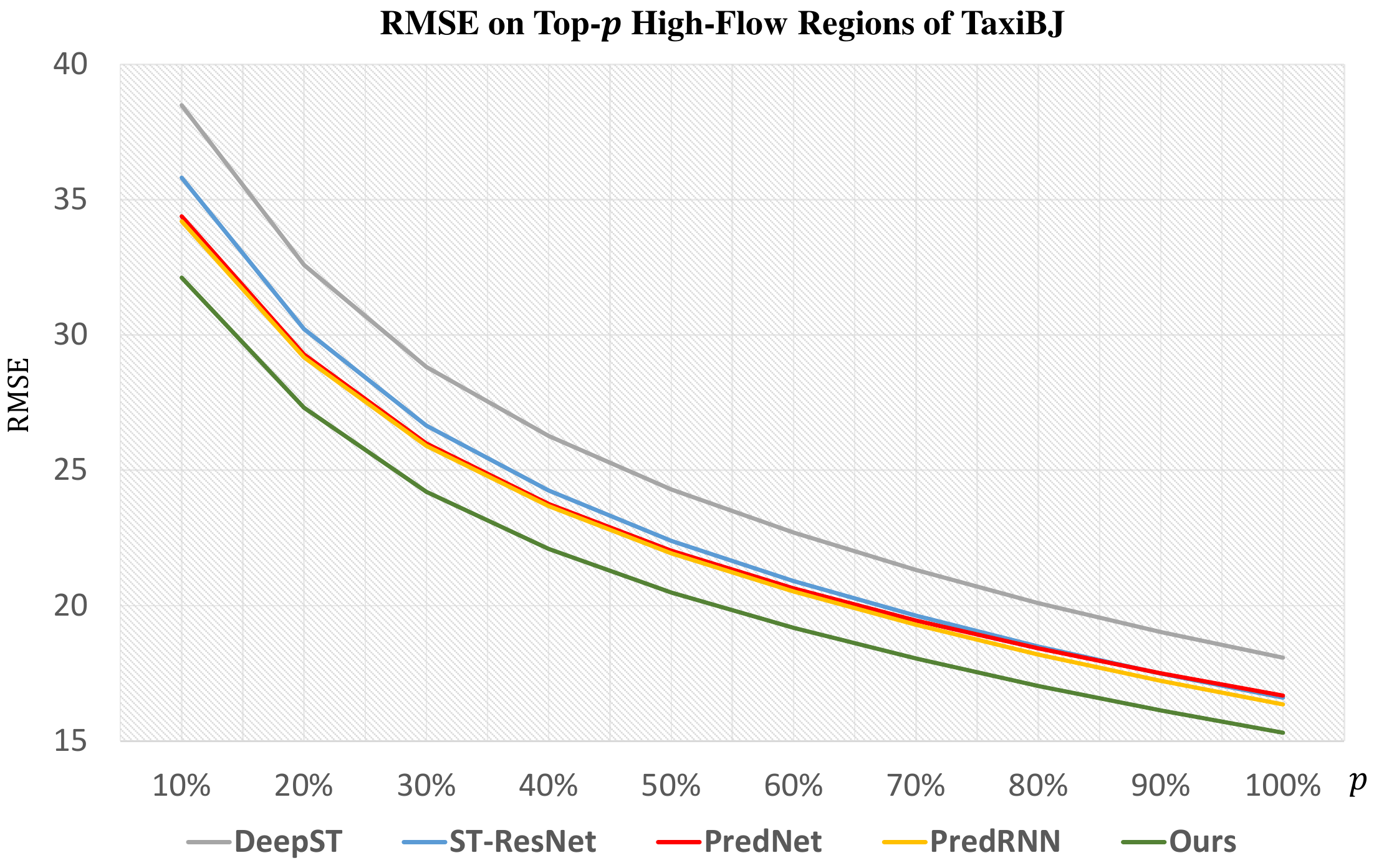}
  \vspace{-10mm}
  \end{center}
   \caption{The RMSE of five deep learning based methods on top-$p$ regions with high traffic flow on the TaxiBJ dataset. $p$ is a percentage. Specifically, we first rank all regions of Beijing on the basis of the average traffic flow and then conduct evaluations on the top-$p$ regions. Best view in color.}
\vspace{0mm}
\label{fig:top-p}
\end{figure}

\begin{figure}[t]
  \begin{center}
     \includegraphics[width=1\columnwidth]{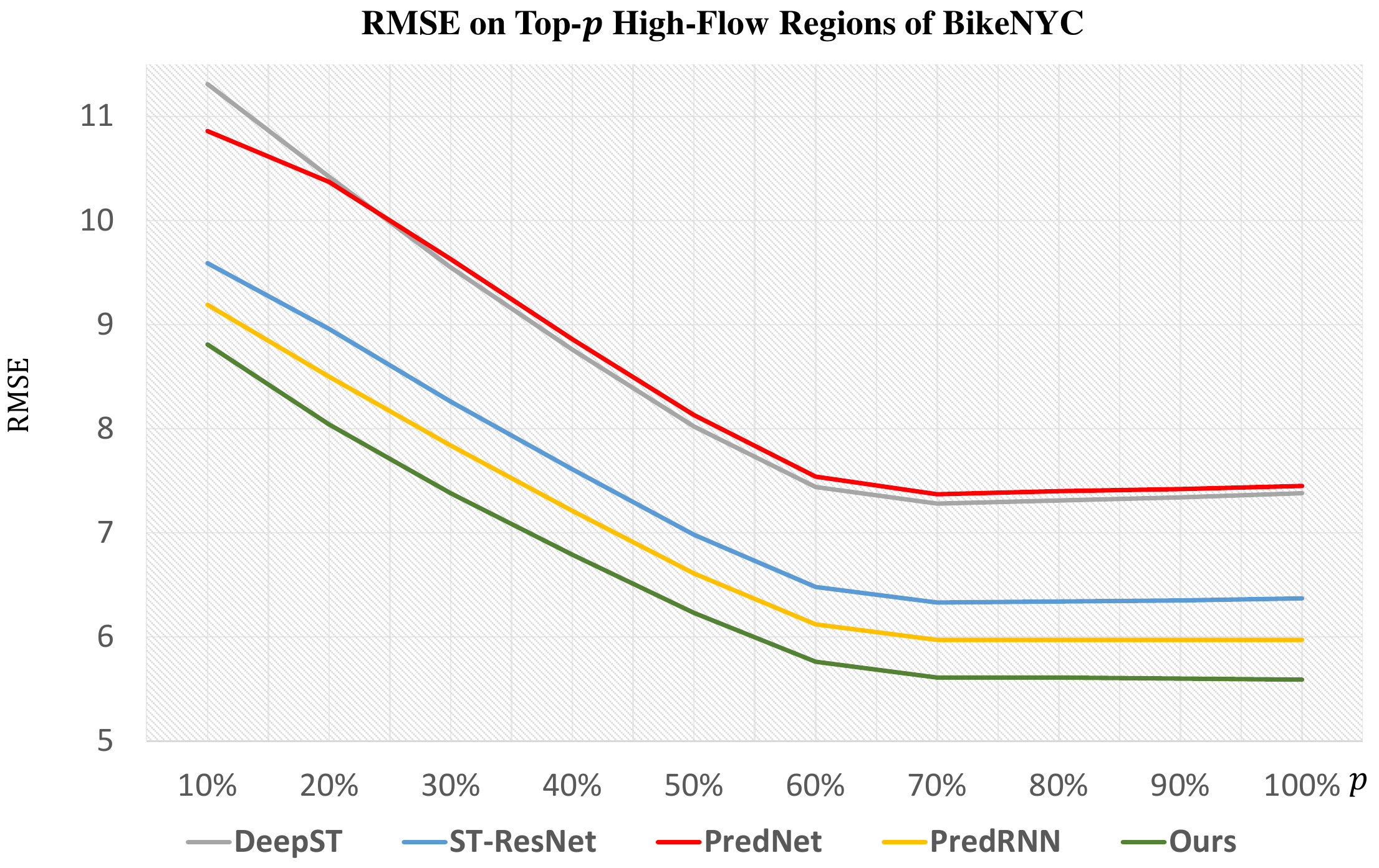}
  \vspace{-10mm}
  \end{center}
   \caption{The RMSE of five deep learning based methods on top-$p$ regions with high traffic flow on the BikeNYC dataset.}
\vspace{0mm}
\label{fig:top-p_NYC}
\end{figure}

\begin{table}
    \caption{Running times of different method on BikeNYC dataset.}
  \vspace{0mm}
\newcommand{\tabincell}[2]{\begin{tabular}{@{}#1@{}}#2\end{tabular}}
  \centering
    \begin{tabular}{c|c}
    \hline
    \tabincell{c}{Model} & \tabincell{c}{Time (ms)}  \\
    \hline\hline
    DeepST & 0.18  \\
    \hline
    ST-ResNet & 2.08   \\
    \hline
    PredNet & 2.71  \\
    \hline
    PredRNN & 4.94 \\
    \hline
    VPN & 12.33  \\
    \hline
    SPN (Ours) & 7.17  \\
    \hline
    \end{tabular}
  \label{tab:time}
\end{table}

{\bf{Comparison on High-Flow Regions:}} {\color{black}Since traffic flow is not uniformly distributed in space, some specific applications are more concerned about the predicted results on congested regions.} In this section, we further measure the RMSE on some regions with high traffic flow. We first rank all regions of Beijing on the basis of the average traffic flow on the training set and then choose the top-$p$ regions ($p$ is a percentage) to conduct the evaluation. As shown in Fig.~\ref{fig:top-p}, {\color{black}on the TaxiBJ dataset,} the RMSE of five deep learning-based methods are much larger on the top-10\% regions and our method obtains a RMSE of 32.11, which shows that this task still has a lot of room for improvement. 
As the percentage $p$ increases, the RMSE of all methods gradually decrease.
{\color{black}As shown in Fig.~\ref{fig:top-p_NYC}, all methods perform poorly on top-10\% regions of BikeNYC, ranging in RMSE from 8.81 to 11.31. As $p$ increases from 10\% to 70\%, their errors gradually decline and no-longer become smaller, since the traffic flow of the remaining 30\% regions is very low.
In summary, our method consistently outperforms other methods under different flow density range $p$ on both TaxiBJ and BikeNYC. These comparisons well demonstrate the superiority of our method. }

{\bf{Efficiency Comparison:}} Finally, we investigate the efficiency of different methods on the TaxiBJ dataset. The running times of six deep learning-based methods are measured with an NVIDIA 1060 GPU. As shown in Table~\ref{tab:time}, DeepST costs 0.18 ms for each inference, while ST-ResNet, PredNet and PredRNN conduct a prediction within 5 ms. Only requiring 7.17 ms, our method is much faster than VPN. In summary, all methods can achieve practical efficiencies. Therefore, the running efficiency is not the bottleneck of this task and we should focus more on the improvement of the performance.

\begin{table}[t]
  \caption{Quantitative comparisons (RMSE) for long-term traffic flow prediction on TaxiBJ. All compared methods have been finetuned for long-term prediction. {\color{black}Each time interval is half an hour (0.5 h) in this dataset.}}
  \vspace{0mm}
\newcommand{\tabincell}[2]{\begin{tabular}{@{}#1@{}}#2\end{tabular}}
  \centering
    \begin{tabular}{c|c|c|c|c}
    \hline
    \multirow{3}{*}{\textbf{Method}} & \multicolumn{4}{c}{\textbf{ Time Interval}} \\
    \cline{2-5}
     & 1 & 2 & 3 & 4\\
     & (0.5 h) & (1.0 h) & (1.5 h) & (2.0 h)\\
     \hline
     \hline
     ST-ResNet & 16.75 & 19.56 & 21.46 & 22.91 \\
     VPN       & 17.42 & 20.50 & 22.58 & 24.26  \\
     PredNet   & 27.55 & 254.68& 255.54& 255.47  \\
     PredRNN   & 16.08 & 19.51 & 20.66 & 22.69  \\
     \hline
     SPN (Ours)       & 15.31 & 19.59 & 23.70 & 28.61  \\
     SPN-LONG (Ours)  & \textbf{\color{red}15.42} & \textbf{\color{red}17.63} & \textbf{\color{red}19.08} & \textbf{\color{red}20.83} \\
    \hline
    \end{tabular}
  \label{tab:long_term_result}
\end{table}

\begin{table}[t]
  \caption{Quantitative comparisons (RMSE) for long-term traffic flow prediction on BikeNYC. {\color{black}Each time interval is an hour (1.0 h) in this dataset.}}
  \vspace{0mm}
\newcommand{\tabincell}[2]{\begin{tabular}{@{}#1@{}}#2\end{tabular}}
  \centering
    \begin{tabular}{c|c|c|c|c}
    \hline
    \multirow{2}{*}{\textbf{Method}} & \multicolumn{4}{c}{\textbf{ Time Interval}} \\
    \cline{2-5}
     & 1 & 2 & 3 & 4\\
     & (1.0 h) & (2.0 h) & (3.0 h) & (4.0 h)\\
     \hline
     \hline
     ST-ResNet & 6.45 & 7.47 & 8.77  & 10.28 \\
     VPN       & 6.55 & 8.01 & 8.86  & 9.41  \\
     PredNet   & 7.46 & 8.95 & 10.08 & 10.93 \\
     PredRNN   & 5.97 & 7.37 & 8.61  & 9.40 \\
     \hline
     SPN (Ours)       & 5.59 & 7.81 & 11.96 & 15.74 \\
     SPN-LONG (Ours)  & \textbf{\color{red}5.81} & \textbf{\color{red}6.80} & \textbf{\color{red}7.54} & \textbf{\color{red}7.90} \\
    \hline
    \end{tabular}
  \label{tab:long_term_BikeNYC_result}
\end{table}

\subsection{Comparison for Long-term Prediction}

In this subsection, we apply the customized SPN-LONG to predict long-term traffic flow and compare it with four deep learning based methods\footnote{On the TaxiBJ dataset, the performances of all compared methods for long-term prediction are directly quoted from \cite{xu2018predcnn}. On the BikeNYC dataset, there is not existing comparison for long-term prediction, thus we implement all compared methods and evaluate their performances.}. These compared methods have been finetuned for long-term prediction.
As shown in Table~\ref{tab:long_term_result}, the RMSE of all methods gradually increases on the TaxiBJ dataset when attempting to forecast the longer-term flow. It can be observed that PredNet performs dreadfully in this scenario, since it was originally designed for single frame prediction and has a low capacity for long-term prediction. By contrast, our method has minor performance degradation and outperforms other methods at each time interval. Specifically, our method achieves the lowest RMSE 20.83 at the fourth time interval and has a relative improvement of 8.2\%, compared with the previous best-performing method PredRNN. Moreover, we also evaluate the original SPN for long-term prediction and it is used to forecast traffic flow in a rolling style. As shown in the penultimate row of Table~\ref{tab:long_term_result}, it performs worse than SPN-LONG, thus we can conclude that it's essential to adapt and retrain SPN for long-term prediction.
We also conduct long-term prediction on BikeNYC dataset. {\color{black}As shown in Table~\ref{tab:long_term_BikeNYC_result}, our SPN-LONG consistently outperforms other compared methods with the best RMSE (e.g., 5.81, 6.80, 7.54 and 7.90 for the $1^{st}$-$4^{th}$ time intervals, respectively). When combining our collected meteorological information, SPN-LONG further reduces the RMSE to 5.72, 6.24, 6.74 and 7.37 for the $1^{st}$-$4^{th}$ intervals, respectively}. These experiments well demonstrate the effectiveness of the customized SPN-LONG for long-term traffic flow prediction.

\begin{figure}[t]
  \begin{center}
     \includegraphics[width=0.9\columnwidth]{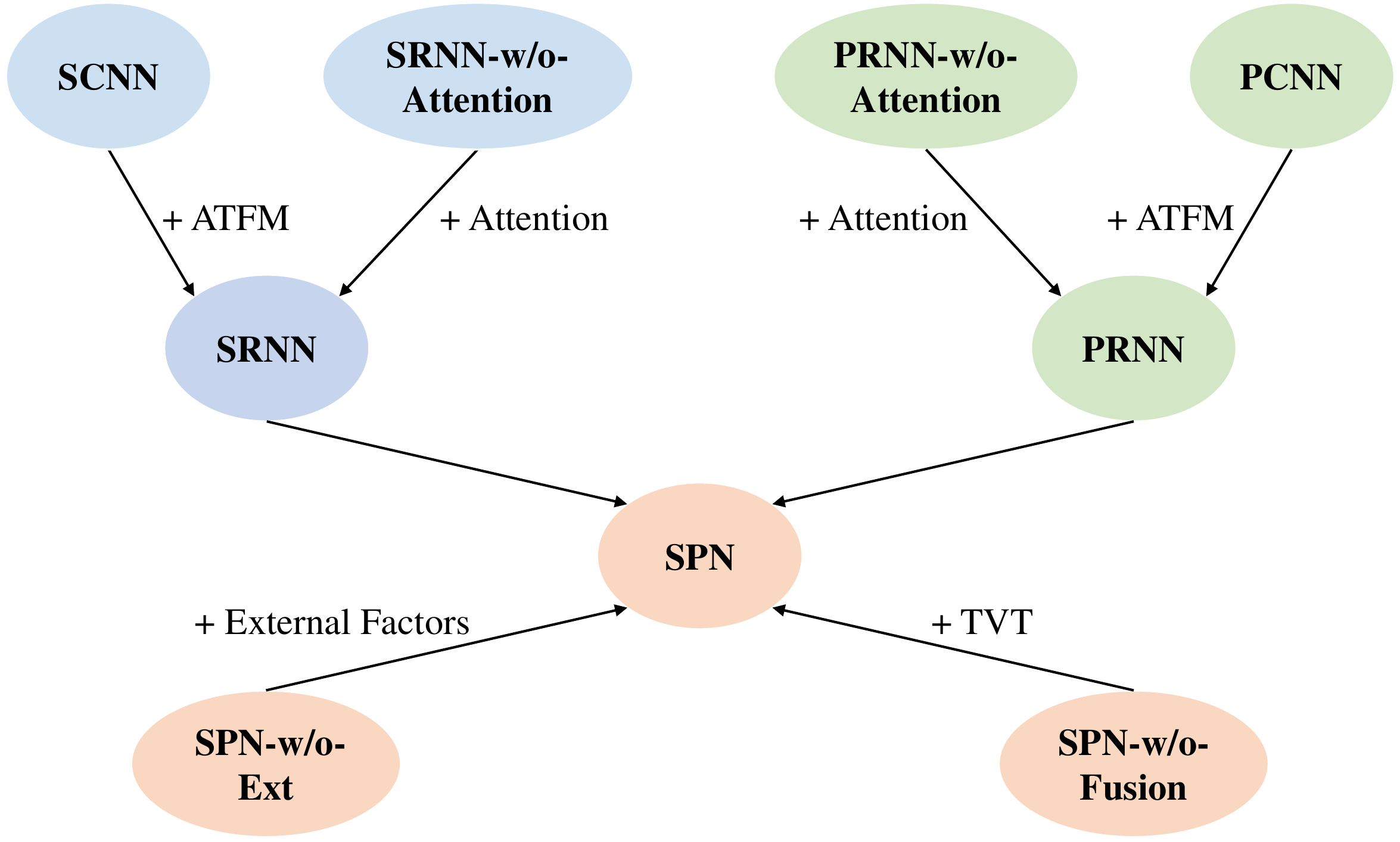}
  \vspace{-4mm}
  \end{center}
   \caption{\color{black}Overview of the differences between all variants of our framework.}
\vspace{0mm}
\label{fig:variants_overview}
\end{figure}

\subsection{Component Analysis}~\label{sec:abl_study}
As described in Section~\ref{sec:framework}, our full model consists of four components: normal feature extraction, sequential representation learning, periodic representation learning and temporally-varying fusion module. In this section, we implement eight variants of our framework in order to verify the effectiveness of each component:
\begin{itemize}
\item \textbf{PCNN:} directly concatenates the periodic features ${P_{in}}$ and feeds them to a convolutional layer with two filters followed by ${tanh}$ for future traffic flow prediction;
\item \textbf{SCNN:} directly concatenates the sequential features ${S_{in}}$ and feeds them to a convolutional layer followed by ${tanh}$ for future traffic flow prediction;
\item \textbf{PRNN-w/o-Attention:} takes periodic features ${P_{in}}$ as input and learns periodic representation with a LSTM layer to predict future traffic flow;
\item \textbf{PRNN:} takes periodic features ${P_{in}}$ as input and learns periodic representation with the proposed ATFM to predict future traffic flow;
\item \textbf{SRNN-w/o-Attention:} takes sequential features ${S_{in}}$ as input and learns sequential representation with a LSTM layer for traffic flow estimation;
\item \textbf{SRNN:} takes sequential features ${S_{in}}$ as input and learns sequential representation with the proposed ATFM to predict future traffic flow;
\item \textbf{SPN-w/o-Ext:} does not consider the effect of external factors and directly trains the model with traffic flow maps;
\item \textbf{SPN-w/o-Fusion:} directly merges sequential representation and periodic representation with equal weight (0.5) to predict future traffic flow.
\end{itemize}

\begin{table}[t]
    \caption{Quantitative comparisons of different variants of our model on the TaxiBJ dataset for component analysis.}
  \vspace{0mm}
\newcommand{\tabincell}[2]{\begin{tabular}{@{}#1@{}}#2\end{tabular}}
  \centering
    \begin{tabular}{c|c|c}
    \hline
    \tabincell{c}{Model} & \tabincell{c}{RMSE} & \tabincell{c}{MAE} \\
    \hline\hline
    PCNN & 33.91 & 17.16 \\
    PRNN-w/o-Attention & 33.51 & 16.70  \\
    PRNN & 32.89 & 16.64 \\
    \hline
    SCNN & 17.15 & 9.56 \\
    SRNN-w/o-Attention & 16.20 & 9.43  \\
    SRNN & 15.82 & 9.34 \\
    \hline
    SPN-w/o-Ext & 16.84 & 9.83\\
    SPN-w/o-Fusion & 15.67 & 9.40\\
    SPN & 15.31 & 9.14  \\
    \hline
    \end{tabular}
  \label{tab:component_results}
\end{table}

{\color{black}The overview of all variants is shown in Fig.~\ref{fig:variants_overview}. First, we use ``SCNN vs. SRNN'' and ``PCNN vs. PRNN'' to verify the effectiveness of ATFM for sequential and periodic representation learning. Then, ``SRNN-w/o-Attention vs SRNN'' and ``PRNN-w/o-Attention vs PRNN'' are conducted to explain the effectiveness of spatial attention. Finally, ``SPN-w/o-Ext vs. SPN'' is utilized to illustrate the influence of external factors and ``SPN-w/o-Fusion vs. SPN'' is utilized to show the effectiveness of Temporally-Varying Fusion (TVF) module.}

\textbf{Effectiveness of ATFM for Sequential Representation Learning:}
As shown in Table~\ref{tab:component_results}, directly concatenating the sequential features ${S}$ for prediction, the baseline variant SCNN gets an RMSE of 17.15. When explicitly modeling the sequential contextual dependencies of traffic flow using the proposed ATFM, SRNN decreases RMSE to 15.82, with 7.75\% relative performance improvement compared to the baseline SCNN, which indicates the effectiveness of the sequential representation learning.

\textbf{Effectiveness of ATFM for Periodic Representation Learning:}
We also explore different network architectures to learn the periodic representation. As shown in Table~\ref{tab:component_results}, the PCNN, which learns to estimate the flow map by simply concatenating all of the periodic features ${P}$, only achieves RMSE of 33.91. In contrast, when introducing ATFM to learn the periodic representation, the RMSE drops to 32.89. This experiment also well demonstrates the effectiveness of the proposed ATFM for spatial-temporal modeling.

\begin{figure*}
\begin{center}
\begin{minipage}[a]{0.925\textwidth}
\includegraphics[width=0.5\textwidth]{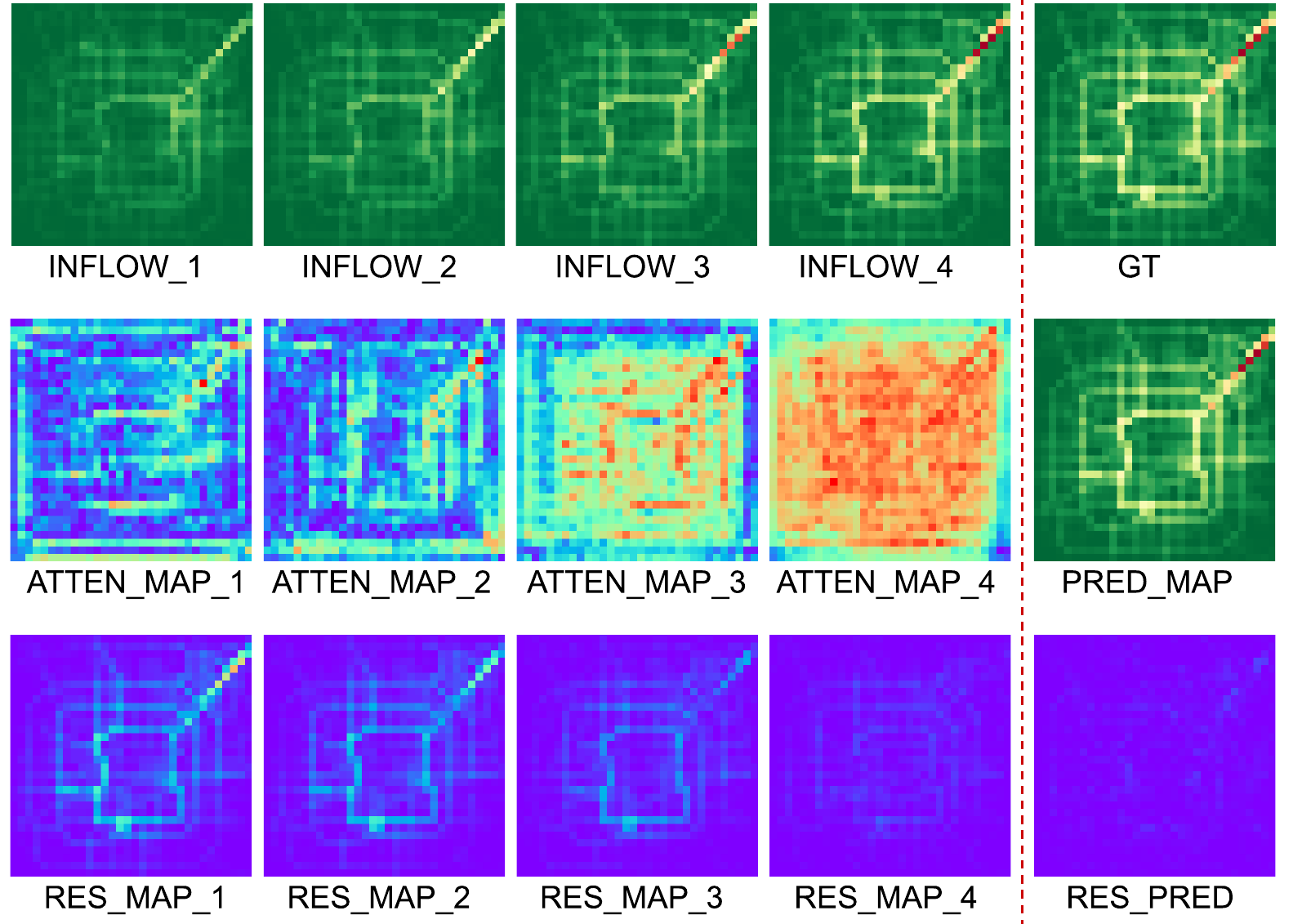}
\includegraphics[width=0.5\textwidth]{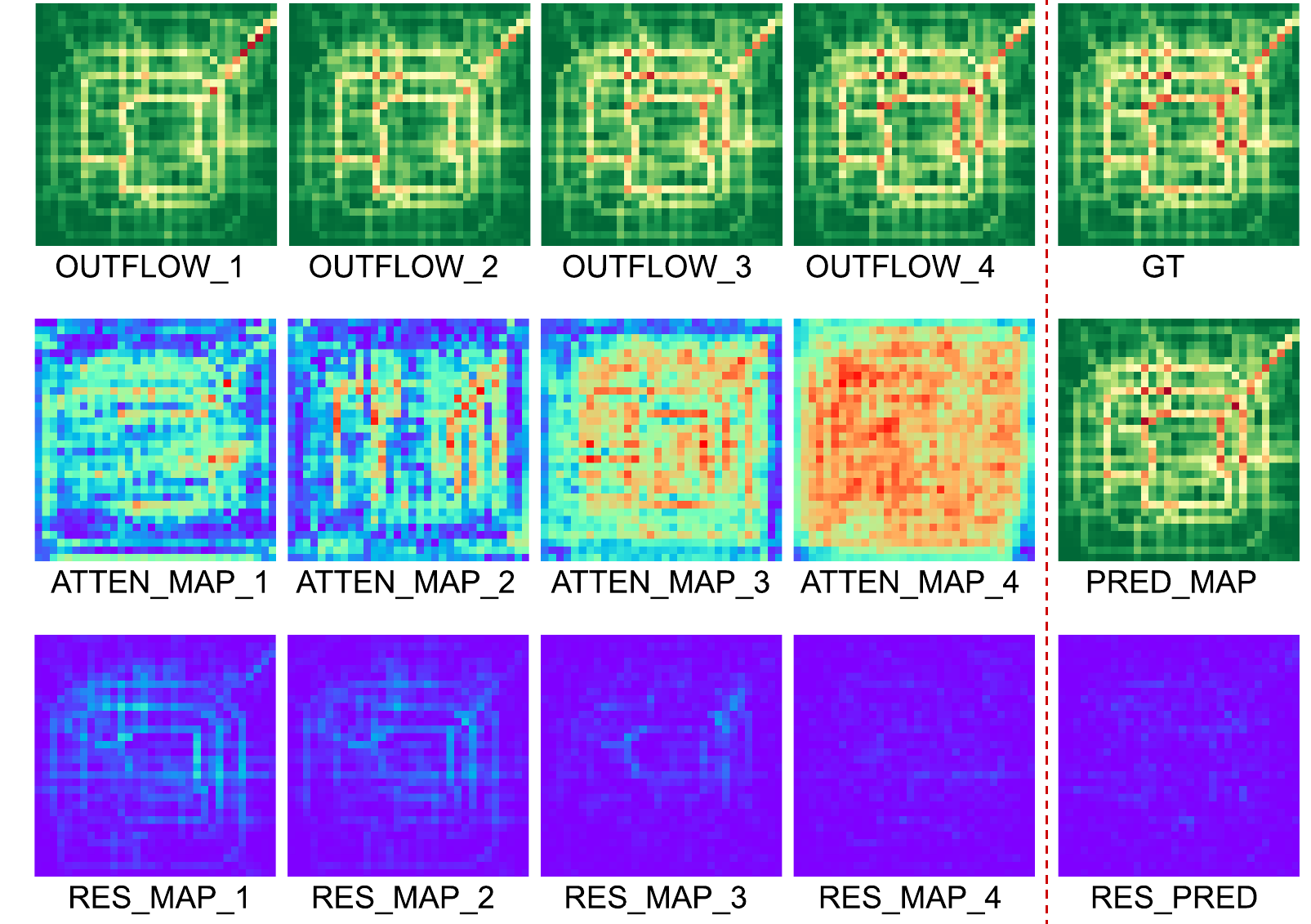}
\end{minipage}
 \vspace{-4mm}
\end{center}
   \caption{Illustration of the generated attention maps of the traffic flow in {\color{black}{sequential representation learning}} with ${n}$ set as 4. Every five columns form one group. In each group:
   i) on the first row, the first four images are the input sequential inflow/outflow maps and the last one is the ground truth inflow/outflow map of next time interval;
   ii) on the second row, the first four images are the attentional maps generated by our ATFM, while the last one is our predicted inflow/outflow map;
   iii) on the third row, the first four images are the residual maps between the input flow maps and the ground truth, while the last one is the residual map between our predicted flow map and the ground truth.
   We can observe that there is a negative correlation between the attentional maps and the residual maps to some extent.
   }
\vspace{-0mm}
\label{fig:Sequential-Attention}
\end{figure*}

\begin{figure*}
\begin{center}
\begin{minipage}[a]{0.775\textwidth}
\includegraphics[width=0.5\textwidth]{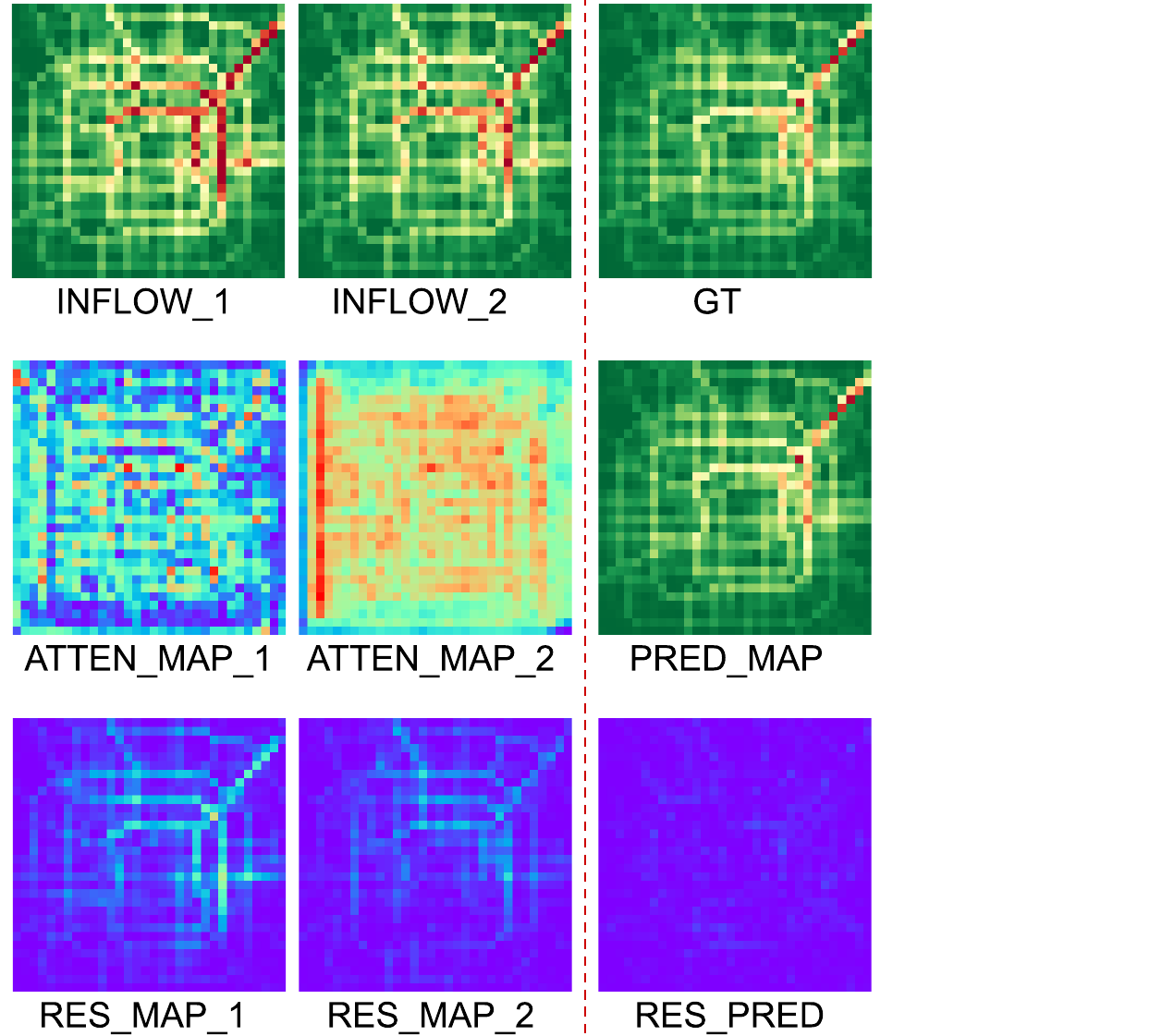}
\includegraphics[width=0.5\textwidth]{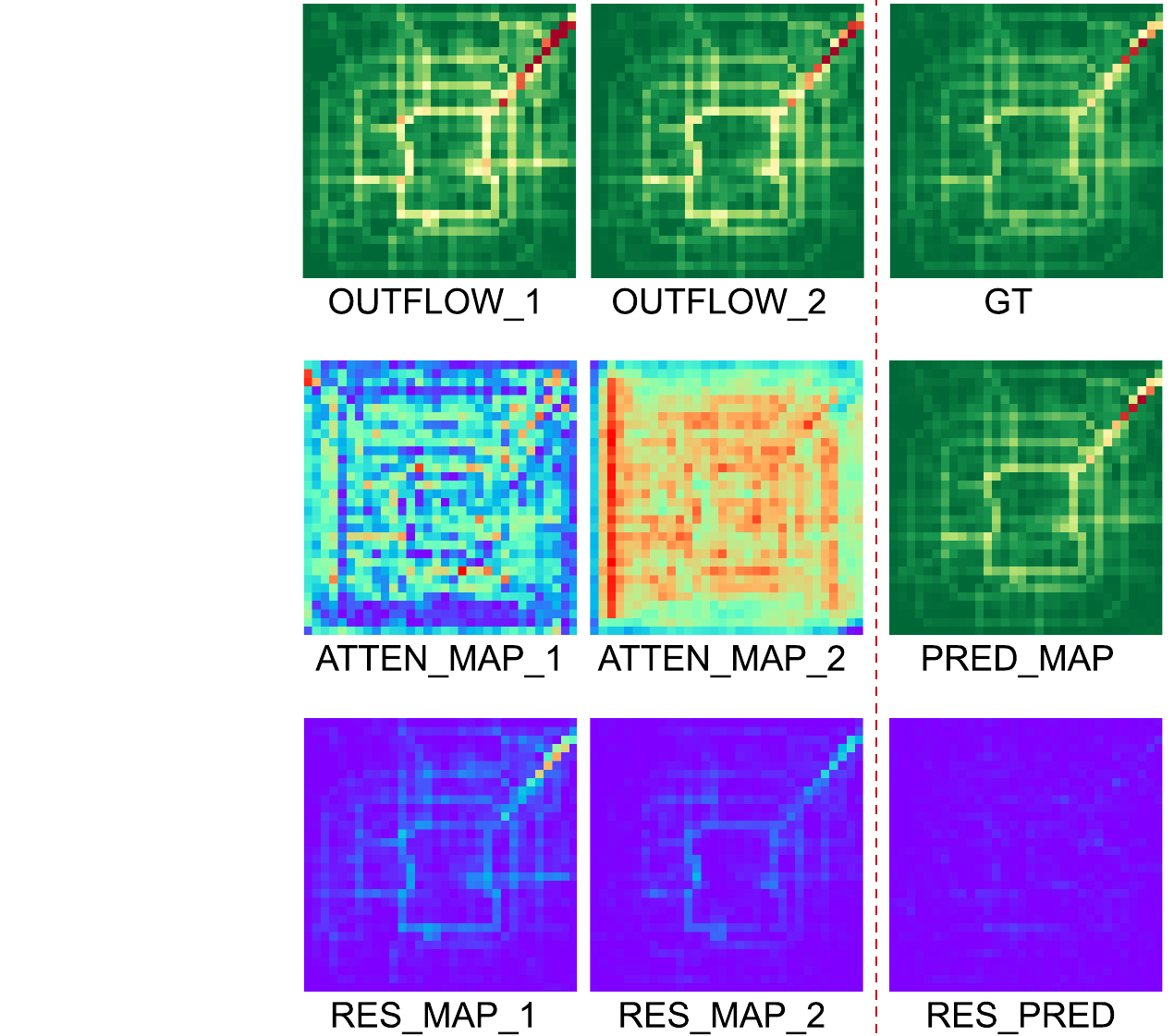}
\end{minipage}
 \vspace{-2mm}
\end{center}
   \caption{Illustration of the generated attentional maps of the traffic flow in {\color{black}{periodic representation learning}} with ${m}$ set as 2. Every three columns form one group. In each group:
   i) on the first row, the first two images are the input periodic inflow/outflow maps and the last one is the ground truth inflow/outflow map of next time interval;
   ii) on the second row, the first two images are the attentional maps generated by our ATFM, while the last one is our predicted inflow/outflow map;
   iii) on the third row, the first two images are the residual maps between the input flow maps and the ground truth, while the last one is the residual map between our predicted flow map and the ground truth.
   We can observe that there is a negative correlation between the attentional maps and the residual maps to some extent.
   }
\vspace{0mm}
\label{fig:Periodic-Attention}
\end{figure*}

\textbf{Effectiveness of Spatial Attention:}
As shown in Table~\ref{tab:component_results}, adopting spatial attention, PRNN decreases the RMSE by 0.62, compared to PRNN-w/o-Attention. For another pair of variants, SRNN with spatial attention has similar performance improvement, compared to SRNN-w/o-Attention. Fig.~\ref{fig:Sequential-Attention} and Fig.~\ref{fig:Periodic-Attention} show some attentional maps generated by our method as well as the residual maps between the input traffic flow maps and their corresponding ground truth. We can observe that there is a negative correlation between the attentional maps and the residual maps to some extent. It indicates that our ATFM is able to capture informative regions at each time step and make better predictions by inferring the trend of evolution. Roughly, the greater difference a region has, the smaller its weight, and vice versa. We can inhibit the impacts of the regions with great differences by multiplying the small weights on their corresponding location features. With the visualization of attentional maps, we can also get to know which regions have the primary positive impacts for the future flow prediction. According to the experiment, we can see that the proposed model can not only effectively improve the prediction accuracy, but also enhance the interpretability of the model to a certain extent.

\textbf{Necessity of External Factors:}
Without modeling the effect of external factors, the variant SPN-w/o-Ext obtains a RMSE of 16.84 on the TaxiBJ dataset and has a performance degradation of 10\%, compared to SPN. The main reason of degradation lies in that some {\color{black}notable} meteorological conditions (e.g., rain and snow) or holidays would seriously affect the traffic flow. Thus, it's necessary to incorporate the external factors to model the traffic flow evolution.

\textbf{Effectiveness of Temporally-Varying Fusion:}
When directly merging the two temporal representations with an equal contribution (0.5), SPN-w/o-fusion achieves a negligible improvement, compared to SRNN. In contrast, after using our proposed fusion strategy, the full model SPN decreases the RMSE from 15.82 to 15.31, with a relative improvement of 3.2\% compared with SRNN. The results show that the contributions of these two representations are not equal and are influenced by various factors. The proposed fusion strategy is effective to adaptively merge the different temporal representations and further improve the performance of traffic flow prediction.

\begin{figure*}
\centering
   \includegraphics[width=1.7\columnwidth]{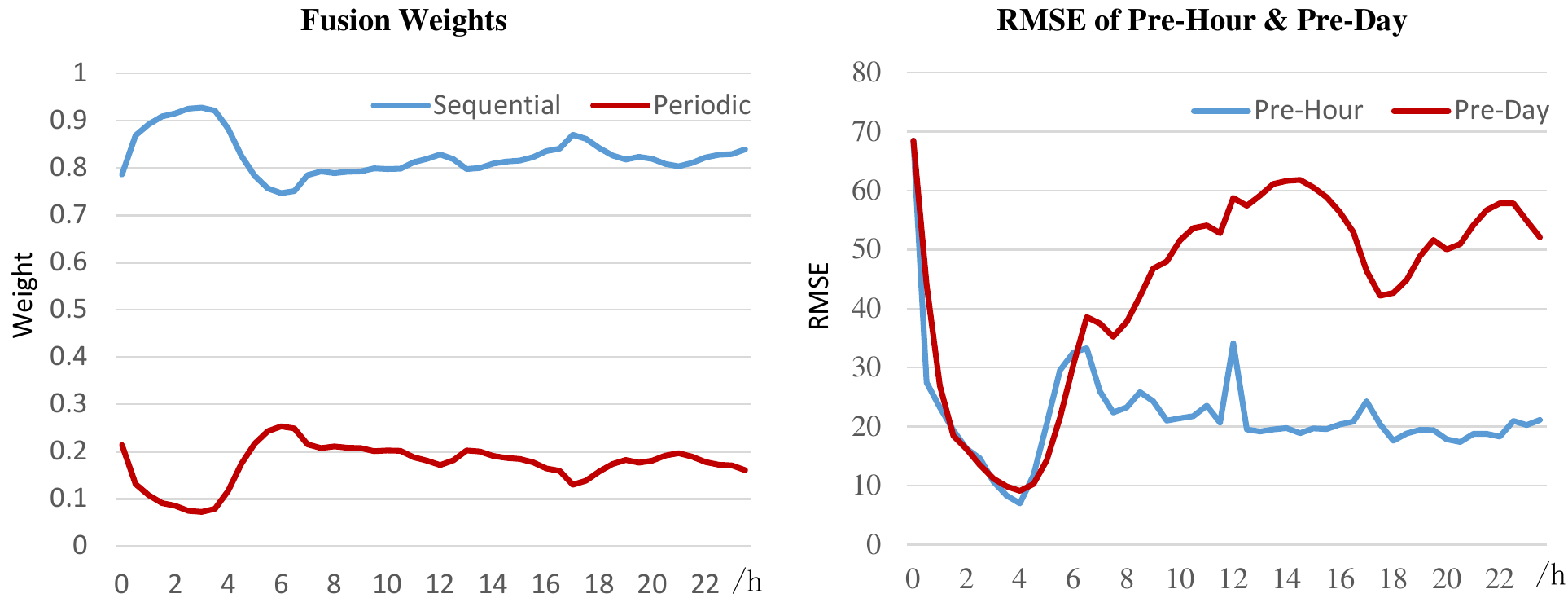}
\vspace{-1mm}
   \caption{\textbf{Left:} The average fusion weights of two types of temporal representation on the testing set of TaxiBJ dataset. \textbf{Right:} The RMSE of traffic flow between two consecutive time intervals (denoted as ``Pre-Hour'') and the RMSE of traffic flow between two adjacent days at the same time interval (denoted as ``Pre-Day'').
   We can find that the weights of sequential representation are greater than that of the periodic representation, which indicates that the sequential trend is more essential for traffic flow prediction.
   }
\vspace{0mm}
\label{fig:ratio_rmse_bj}
\end{figure*}

\textbf{Further Discussion:}
To analyze how each temporal representation contributes to the performance of traffic flow prediction, we measure the average fusion weights of two temporal representations at each time interval on the testing set. As shown in the left of Fig.~\ref{fig:ratio_rmse_bj}, the fusion weights of sequential representation are greater than that of the periodic representation. To explain this phenomenon, we further measure \textbf{i)} the RMSE of traffic flow between two consecutive time intervals, denoted as ``Pre-Hour'', and \textbf{ii)} the RMSE of traffic flow between two adjacent days at the same time interval, denoted as ``Pre-Day''. As shown on the right of Fig.~\ref{fig:ratio_rmse_bj}, the RMSE of
``Pre-Day'' is much higher than that of ``Pre-Hour'' at most time except for the wee hours. Based on this observation, we can conclude that the sequential representation is more essential for the traffic flow prediction, since the sequential data is more regular.
Although the weight is low, the periodic representation still helps to improve the performance of traffic flow prediction qualitatively and quantitatively. For example, we can decrease the RMSE of SRNN by 3.2\% after incorporating the periodic representation. 

\subsection{Extension to Citywide Passenger Demand Prediction}\label{sec:demand}

Our ATFM is a general model for urban mobility modeling. Apart from the traffic flow prediction, it can also be applied to other related traffic tasks, such as citywide passenger demand prediction. In this subsection, we extend the proposed method to forecast the passenger pickup/dropoff demands at the next time interval (half an hour) with historical mobility trips.

We conduct experiments with taxi trips in New York City. Since most taxi transactions were made in the Manhattan borough, we choose it as the studied area and divide it into a $h{\times}w$ grid map. We collect 132 million taxicab trip records during 2014 from New York City Taxi and Limousine Commission (NYCTLC\footnote{https://www1.nyc.gov/site/tlc/about/tlc-trip-record-data.page}). Each record contains the timestamp and the geo-coordinates of pickup and dropoff locations. For each region, we measure the passenger pickup/dropoff demands every half an hour, thus the dimensionality of passenger demand maps is $2{\times}h{\times}w$. We collect external meteorological factors (e.g., temperature, wind speed and weather conditions) from Wunderground and the holidays are also marked. Finally, we train our model with the historical demand of the first 337 days and conduct evaluation with the data in the last four weeks. 

\begin{table}[t]
  \caption{Effectiveness of different spatial resolutions for short-term demand prediction.}
  \vspace{0mm}
\newcommand{\tabincell}[2]{\begin{tabular}{@{}#1@{}}#2\end{tabular}}
  \centering
    \begin{tabular}{c|c|c}
    \hline
    Spatial Resolution & RMSE & MAE\\
    \hline
    \hline
    $10\times3$ & 32.28 & 19.24 \\
    \hline
    $12\times4$ & 23.46 & 13.63 \\
    \hline
    $15\times5$ & 17.29 & 9.91 \\
    \hline
    $20\times7$ & 11.80 & 6.50 \\
    \hline
    $30\times10$ & 7.38 & 3.96 \\
    \hline
    \end{tabular}

  \label{tab:resolution}
\end{table}

\begin{table}[t]
  \caption{Quantitative comparisons for citywide passenger short-term demand prediction.}
  \vspace{0mm}
\newcommand{\tabincell}[2]{\begin{tabular}{@{}#1@{}}#2\end{tabular}}
  \centering
    \begin{tabular}{c|c|c}
    \hline
    Method & RMSE & MAE\\
    \hline
    \hline
    HA & 39.02 & 20.24 \\
    \hline
    VPN & 18.70 & 10.60 \\
    \hline
    DeepST & 18.55 & 10.77 \\
    \hline
    ST-ResNet & 18.20 & 10.14 \\
    \hline
    PredNet & 18.53 & 11.01 \\
    \hline
    PredRNN & 17.82 & 10.34 \\
    \hline
    SPN (Ours) & \textbf{\color{red}17.29} & \textbf{\color{red}9.91} \\
    \hline
    \end{tabular}

  \label{tab:passenger_demand}
\end{table}

\begin{table}[t]
  \caption{Quantitative comparisons (RMSE) for citywide passenger long-term demand prediction. All compared methods have been finetuned for long-term prediction. {\color{black}Each time interval is half an hour (0.5 h) in this dataset.}}
  \vspace{0mm}
\newcommand{\tabincell}[2]{\begin{tabular}{@{}#1@{}}#2\end{tabular}}
  \centering
    \begin{tabular}{c|c|c|c|c}
    \hline
    \multirow{3}{*}{\textbf{Method}} & \multicolumn{4}{c}{\textbf{ Time Interval}} \\
    \cline{2-5}
     & 1 & 2 & 3 & 4\\
     & (0.5 h) & (1.0 h) & (1.5 h) & (2.0 h)\\
     \hline
     \hline
     ST-ResNet & 18.11 & 22.87 & 28.21 & 34.51 \\
     VPN       & 19.74 & 22.63 & 25.36 & 28.19 \\
     PredNet   & 18.44 & 22.44 & 25.97 & 29.34 \\
     PredRNN   & 17.75 & 21.62 & 25.41 & 29.31 \\
     \hline
     SPN-LONG (Ours)  & \textbf{\color{red}17.41} & \textbf{\color{red}20.08} & \textbf{\color{red}22.19} & \textbf{\color{red}24.45} \\
    \hline
    \end{tabular}
  \label{tab:long_term_demand}
\end{table}

We first explore the effectiveness of different spatial resolutions ($h{\times}w$). As shown in Table~\ref{tab:resolution}, the RMSE and MAE of our method gradually decrease as the resolution increases. However, this performance improvement may come from the corresponding reduction in demand as the unit area becomes smaller. 
 Moreover, too high resolution may result in over-divided regions (e.g., a stadium may be divided into multi regions) and it is unnecessary to forecast taxi demand in a very small region. In the previous work~\cite{yao2018deep}, Didi Chuxing, a famous taxi requesting company in China, predicted taxi demand in each $0.7km \times 0.7km$ region. Following this setting, we divide the Manhattan borough into a $15{\times}7$ grid map and each grid represents a geographical region with a size of about $0.75km \times 0.75km$.

We then compare our method with HA and five deep learning based methods. As shown in Table~\ref{tab:passenger_demand}, the baseline method HA obtains a RMSE of 39.02 and a MAE of 20.24, which is impractical in the taxi industry. By contrast, our method dramatically decreases the RMSE to 17.29 and outperforms other compared methods for short-term prediction.
Moreover, we adapt and retrain these deep learning based methods to forecast the long-term demand and summarize their RMSE in Table~\ref{tab:long_term_demand}. It can be observed that our SPN-LONG model achieves the best performance at every time interval. In particular, our method has a performance improvement of 16.58\% compared with PredRNN at the fourth time interval.
These experiments show that the proposed method is also effective for passenger demand prediction.

\section{Conclusion}~\label{sec:conclusion}
In this work, we utilize massive human trajectory data collected from mobility digital devices to study the traffic flow prediction problem. Its key challenge lies in how to adaptively integrate various factors that affect the flow changes, such as sequential trends, periodic laws and spatial dependencies.
To address these issues, we propose a novel Attentive Traffic Flow Machine (ATFM), which explicitly learns dynamic spatial-temporal representations from historical traffic flow maps with an attention mechanism. Based on the proposed ATFM, we develop a unified framework to adaptively merge the sequential and periodic representations with the aid of a temporally-varying fusion module for citywide traffic flow prediction.
By conducting extensive experiments on two public benchmarks, we have verified the effectiveness of our method for traffic flow prediction. Moreover, to verify the generalization of ATFM, we apply the customized framework to forecast the passenger pickup/dropoff demand and it can also achieve practical performance on this traffic prediction task.

However, there is still much room for improvement. First, it may be suboptimal to divide the studied cities into regular grid maps. In future work, we would divide them into traffic analysis zones with irregular shapes on the basis of the functionalities of regions. We would model such traffic systems as graphs and adapt Graph Convolutional Network (GCN~\cite{duvenaud2015convolutional,chen2020physical}) to learn spatial-temporal features. 
Second, the functionality information of zones has not been fully explored in most previous works. Intuitively, the zones with the same functionalities usually have similar traffic flow patterns. For instance, most residential regions have high outflow during morning rush hours and have high inflow during evening rush hours. Base on this consideration, we plan to incorporate the prior knowledge of functionality information of zones (e.g., the Point of Interest (POI) data, land-use data and socio-demographic data) into GCN to further improve the prediction accuracy. 


\ifCLASSOPTIONcaptionsoff
  \newpage
\fi

\bibliographystyle{IEEEtran}
\bibliography{ATFM-Reference}

\begin{thebibliography}{10}
\providecommand{\url}[1]{#1}
\csname url@samestyle\endcsname
\providecommand{\newblock}{\relax}
\providecommand{\bibinfo}[2]{#2}
\providecommand{\BIBentrySTDinterwordspacing}{\spaceskip=0pt\relax}
\providecommand{\BIBentryALTinterwordstretchfactor}{4}
\providecommand{\BIBentryALTinterwordspacing}{\spaceskip=\fontdimen2\font plus
\BIBentryALTinterwordstretchfactor\fontdimen3\font minus
  \fontdimen4\font\relax}
\providecommand{\BIBforeignlanguage}[2]{{%
\expandafter\ifx\csname l@#1\endcsname\relax
\typeout{** WARNING: IEEEtran.bst: No hyphenation pattern has been}%
\typeout{** loaded for the language `#1'. Using the pattern for}%
\typeout{** the default language instead.}%
\else
\language=\csname l@#1\endcsname
\fi
#2}}
\providecommand{\BIBdecl}{\relax}
\BIBdecl

\bibitem{xingjian2015convolutional}
S.~Xingjian, Z.~Chen, H.~Wang, D.-Y. Yeung, W.-K. Wong, and W.-c. Woo,
  ``Convolutional lstm network: A machine learning approach for precipitation
  nowcasting,'' in \emph{NIPS}, 2015, pp. 802--810.

\bibitem{zheng2014urban}
Y.~Zheng, L.~Capra, O.~Wolfson, and H.~Yang, ``Urban computing: concepts,
  methodologies, and applications,'' \emph{TIST}, vol.~5, no.~3, p.~38, 2014.

\bibitem{zhang2011data}
J.~Zhang, F.-Y. Wang, K.~Wang, W.-H. Lin, X.~Xu, and C.~Chen, ``Data-driven
  intelligent transportation systems: A survey,'' \emph{IEEE Transactions on
  Intelligent Transportation Systems}, vol.~12, no.~4, pp. 1624--1639, 2011.

\bibitem{huang2014deep}
W.~Huang, G.~Song, H.~Hong, and K.~Xie, ``Deep architecture for traffic flow
  prediction: deep belief networks with multitask learning,'' \emph{IEEE
  Transactions on Intelligent Transportation Systems}, vol.~15, no.~5, pp.
  2191--2201, 2014.

\bibitem{lv2014traffic}
Y.~Lv, Y.~Duan, W.~Kang, Z.~Li, and F.-Y. Wang, ``Traffic flow prediction with
  big data: a deep learning approach,'' \emph{IEEE Transactions on Intelligent
  Transportation Systems}, vol.~16, no.~2, pp. 865--873, 2014.

\bibitem{polson2017deep}
N.~G. Polson and V.~O. Sokolov, ``Deep learning for short-term traffic flow
  prediction,'' \emph{Transportation Research Part C: Emerging Technologies},
  vol.~79, pp. 1--17, 2017.

\bibitem{zhang2017deep}
J.~Zhang, Y.~Zheng, and D.~Qi, ``Deep spatio-temporal residual networks for
  citywide crowd flows prediction.'' in \emph{AAAI}, 2017, pp. 1655--1661.

\bibitem{shekhar2007adaptive}
S.~Shekhar and B.~M. Williams, ``Adaptive seasonal time series models for
  forecasting short-term traffic flow,'' \emph{Transportation Research Record},
  vol. 2024, no.~1, pp. 116--125, 2007.

\bibitem{guo2014adaptive}
J.~Guo, W.~Huang, and B.~M. Williams, ``Adaptive kalman filter approach for
  stochastic short-term traffic flow rate prediction and uncertainty
  quantification,'' \emph{Transportation Research Part C: Emerging
  Technologies}, vol.~43, pp. 50--64, 2014.

\bibitem{zheng2013time}
J.~Zheng and L.~M. Ni, ``Time-dependent trajectory regression on road networks
  via multi-task learning,'' in \emph{AAAI}, 2013, pp. 1048--1055.

\bibitem{deng2016latent}
D.~Deng, C.~Shahabi, U.~Demiryurek, L.~Zhu, R.~Yu, and Y.~Liu, ``Latent space
  model for road networks to predict time-varying traffic,'' \emph{KDD}, pp.
  1525--1534, 2016.

\bibitem{zhang2016dnn}
J.~Zhang, Y.~Zheng, D.~Qi, R.~Li, and X.~Yi, ``Dnn-based prediction model for
  spatio-temporal data,'' in \emph{SIGSPATIAL}.\hskip 1em plus 0.5em minus
  0.4em\relax ACM, 2016, p.~92.

\bibitem{xu2018predcnn}
Z.~Xu, Y.~Wang, M.~Long, J.~Wang, and M.~KLiss, ``Predcnn: Predictive learning
  with cascade convolutions.'' in \emph{IJCAI}, 2018, pp. 2940--2947.

\bibitem{zhang2019flow}
J.~Zhang, Y.~Zheng, J.~Sun, and D.~Qi, ``Flow prediction in spatio-temporal
  networks based on multitask deep learning,'' \emph{TKDE}, 2019.

\bibitem{zhao2017lstm}
Z.~Zhao, W.~Chen, X.~Wu, P.~C. Chen, and J.~Liu, ``Lstm network: a deep
  learning approach for short-term traffic forecast,'' \emph{IET Intelligent
  Transport Systems}, vol.~11, no.~2, pp. 68--75, 2017.

\bibitem{zhang2018predicting}
J.~Zhang, Y.~Zheng, D.~Qi, R.~Li, X.~Yi, and T.~Li, ``Predicting citywide crowd
  flows using deep spatio-temporal residual networks,'' \emph{Artificial
  Intelligence}, vol. 259, pp. 147--166, 2018.

\bibitem{sharma2015action}
S.~Sharma, R.~Kiros, and R.~Salakhutdinov, ``Action recognition using visual
  attention,'' \emph{arXiv:1511.04119}, 2015.

\bibitem{lu2016knowing}
J.~Lu, C.~Xiong, D.~Parikh, and R.~Socher, ``Knowing when to look: Adaptive
  attention via a visual sentinel for image captioning,''
  \emph{arXiv:1612.01887}, 2016.

\bibitem{liu2018crowd}
L.~Liu, H.~Wang, G.~Li, W.~Ouyang, and L.~Lin, ``Crowd counting using deep
  recurrent spatial-aware network,'' in \emph{IJCAI}.\hskip 1em plus 0.5em
  minus 0.4em\relax AAAI Press, 2018, pp. 849--855.

\bibitem{tay2019aanet}
C.-P. Tay, S.~Roy, and K.-H. Yap, ``Aanet: Attribute attention network for
  person re-identifications,'' in \emph{CVPR}, 2019, pp. 7134--7143.

\bibitem{liu2018attentive}
L.~Liu, R.~Zhang, J.~Peng, G.~Li, B.~Du, and L.~Lin, ``Attentive crowd flow
  machines,'' in \emph{2018 ACM Multimedia Conference on Multimedia
  Conference}.\hskip 1em plus 0.5em minus 0.4em\relax ACM, 2018, pp.
  1553--1561.

\bibitem{williams1998urban}
B.~Williams, P.~Durvasula, and D.~Brown, ``Urban freeway traffic flow
  prediction: application of seasonal autoregressive integrated moving average
  and exponential smoothing models,'' \emph{Transportation Research Record:
  Journal of the Transportation Research Board}, no. 1644, pp. 132--141, 1998.

\bibitem{castro2009online}
M.~Castro-Neto, Y.-S. Jeong, M.-K. Jeong, and L.~D. Han, ``Online-svr for
  short-term traffic flow prediction under typical and atypical traffic
  conditions,'' \emph{Expert systems with applications}, vol.~36, no.~3, pp.
  6164--6173, 2009.

\bibitem{li2012prediction}
X.~Li, G.~Pan, Z.~Wu, G.~Qi, S.~Li, D.~Zhang, W.~Zhang, and Z.~Wang,
  ``Prediction of urban human mobility using large-scale taxi traces and its
  applications,'' \emph{Frontiers of Computer Science}, vol.~6, no.~1, pp.
  111--121, 2012.

\bibitem{lippi2013short}
M.~Lippi, M.~Bertini, and P.~Frasconi, ``Short-term traffic flow forecasting:
  An experimental comparison of time-series analysis and supervised learning,''
  \emph{IEEE Transactions on Intelligent Transportation Systems}, vol.~14,
  no.~2, pp. 871--882, 2013.

\bibitem{johansen1991estimation}
S.~Johansen, ``Estimation and hypothesis testing of cointegration vectors in
  gaussian vector autoregressive models,'' \emph{Econometrica: journal of the
  Econometric Society}, pp. 1551--1580, 1991.

\bibitem{box2015time}
G.~E. Box, G.~M. Jenkins, G.~C. Reinsel, and G.~M. Ljung, \emph{Time series
  analysis: forecasting and control}, 2015.

\bibitem{duan2016efficient}
Y.~Duan, Y.~Lv, Y.-L. Liu, and F.-Y. Wang, ``An efficient realization of deep
  learning for traffic data imputation,'' \emph{Transportation research part C:
  emerging technologies}, vol.~72, pp. 168--181, 2016.

\bibitem{chen2017visual}
Z.~Chen, J.~Zhou, and X.~Wang, ``Visual analytics of movement pattern based on
  time-spatial data: A neural net approach,'' \emph{arXiv preprint
  arXiv:1707.02554}, 2017.

\bibitem{fouladgar2017scalable}
M.~Fouladgar, M.~Parchami, R.~Elmasri, and A.~Ghaderi, ``Scalable deep traffic
  flow neural networks for urban traffic congestion prediction,'' \emph{arXiv
  preprint arXiv:1703.01006}, 2017.

\bibitem{ke2017short}
J.~Ke, H.~Zheng, H.~Yang, and X.~M. Chen, ``Short-term forecasting of passenger
  demand under on-demand ride services: A spatio-temporal deep learning
  approach,'' \emph{Transportation Research Part C: Emerging Technologies},
  vol.~85, pp. 591--608, 2017.

\bibitem{wei2018intellilight}
H.~Wei, G.~Zheng, H.~Yao, and Z.~Li, ``Intellilight: A reinforcement learning
  approach for intelligent traffic light control,'' in \emph{KDD}.\hskip 1em
  plus 0.5em minus 0.4em\relax ACM, 2018, pp. 2496--2505.

\bibitem{liu2019crowd}
L.~Liu, Z.~Qiu, G.~Li, S.~Liu, W.~Ouyang, and L.~Lin, ``Crowd counting with
  deep structured scale integration network,'' in \emph{ICCV}, 2019, pp.
  1774--1783.

\bibitem{he2016deep}
K.~He, X.~Zhang, S.~Ren, and J.~Sun, ``Deep residual learning for image
  recognition,'' in \emph{CVPR}, 2016, pp. 770--778.

\bibitem{geng2019spatiotemporal}
X.~Geng, Y.~Li, L.~Wang, L.~Zhang, Q.~Yang, J.~Ye, and Y.~Liu, ``Spatiotemporal
  multi-graph convolution network for ride-hailing demand forecasting,'' in
  \emph{AAAI}, 2019.

\bibitem{wang2018crowd}
L.~Wang, X.~Geng, X.~Ma, F.~Liu, and Q.~Yang, ``Crowd flow prediction by deep
  spatio-temporal transfer learning,'' \emph{arXiv preprint arXiv:1802.00386},
  2018.

\bibitem{yao2019learning}
H.~Yao, Y.~Liu, Y.~Wei, X.~Tang, and Z.~Li, ``Learning from multiple cities: A
  meta-learning approach for spatial-temporal prediction,'' \emph{arXiv
  preprint arXiv:1901.08518}, 2019.

\bibitem{luong2015effective}
M.-T. Luong, H.~Pham, and C.~D. Manning, ``Effective approaches to
  attention-based neural machine translation,'' \emph{arXiv:1508.04025}, 2015.

\bibitem{graves2013speech}
A.~Graves, A.-r. Mohamed, and G.~Hinton, ``Speech recognition with deep
  recurrent neural networks,'' in \emph{ICASSP}.\hskip 1em plus 0.5em minus
  0.4em\relax IEEE, 2013, pp. 6645--6649.

\bibitem{kalchbrenner2017video}
N.~Kalchbrenner, A.~van~den Oord, K.~Simonyan, I.~Danihelka, O.~Vinyals,
  A.~Graves, and K.~Kavukcuoglu, ``Video pixel networks,'' in \emph{ICML},
  2017, pp. 1771--1779.

\bibitem{lotter2017deep}
W.~Lotter, G.~Kreiman, and D.~Cox, ``Deep predictive coding networks for video
  prediction and unsupervised learning,'' in \emph{ICLR}, 2017.

\bibitem{wang2017predrnn}
Y.~Wang, M.~Long, J.~Wang, Z.~Gao, and S.~Y. Philip, ``Predrnn: Recurrent
  neural networks for predictive learning using spatiotemporal lstms,'' in
  \emph{NIPS}, 2017, pp. 879--888.

\bibitem{tian2015predicting}
Y.~Tian and L.~Pan, ``Predicting short-term traffic flow by long short-term
  memory recurrent neural network,'' in \emph{2015 IEEE international
  conference on smart city/SocialCom/SustainCom (SmartCity)}.\hskip 1em plus
  0.5em minus 0.4em\relax IEEE, 2015, pp. 153--158.

\bibitem{fu2016using}
R.~Fu, Z.~Zhang, and L.~Li, ``Using lstm and gru neural network methods for
  traffic flow prediction,'' in \emph{2016 31st Youth Academic Annual
  Conference of Chinese Association of Automation (YAC)}.\hskip 1em plus 0.5em
  minus 0.4em\relax IEEE, 2016, pp. 324--328.

\bibitem{mackenzie2018evaluation}
J.~Mackenzie, J.~F. Roddick, and R.~Zito, ``An evaluation of htm and lstm for
  short-term arterial traffic flow prediction,'' \emph{IEEE Transactions on
  Intelligent Transportation Systems}, no.~99, pp. 1--11, 2018.

\bibitem{xu2016ask}
H.~Xu and K.~Saenko, ``Ask, attend and answer: Exploring question-guided
  spatial attention for visual question answering,'' in \emph{ECCV}.\hskip 1em
  plus 0.5em minus 0.4em\relax Springer, 2016, pp. 451--466.

\bibitem{liu2019contextualized}
L.~Liu, Z.~Qiu, G.~Li, Q.~Wang, Ouyang，Wanli, and L.~Lin, ``Contextualized
  spatial-temporal network for taxi origin-destination demand prediction,''
  \emph{IEEE Transactions on Intelligent Transportation Systems}, 2019.

\bibitem{harris2010digital}
D.~Harris and S.~Harris, \emph{Digital design and computer architecture}.\hskip
  1em plus 0.5em minus 0.4em\relax Morgan Kaufmann, 2010.

\bibitem{paszke2017automatic}
A.~Paszke, S.~Gross, S.~Chintala, G.~Chanan, E.~Yang, Z.~DeVito, Z.~Lin,
  A.~Desmaison, L.~Antiga, and A.~Lerer, ``Automatic differentiation in
  pytorch,'' in \emph{NIPS workshop}, 2017.

\bibitem{glorot2010understanding}
X.~Glorot and Y.~Bengio, ``Understanding the difficulty of training deep
  feedforward neural networks,'' in \emph{AISTATS}, 2010, pp. 249--256.

\bibitem{kingma2014adam}
D.~Kingma and J.~Ba, ``Adam: A method for stochastic optimization,''
  \emph{arXiv:1412.6980}, 2014.

\bibitem{yao2018deep}
H.~Yao, F.~Wu, J.~Ke, X.~Tang, Y.~Jia, S.~Lu, P.~Gong, and J.~Ye, ``Deep
  multi-view spatial-temporal network for taxi demand prediction,'' \emph{arXiv
  preprint arXiv:1802.08714}, 2018.

\bibitem{duvenaud2015convolutional}
D.~K. Duvenaud, D.~Maclaurin, J.~Iparraguirre, R.~Bombarell, T.~Hirzel,
  A.~Aspuru-Guzik, and R.~P. Adams, ``Convolutional networks on graphs for
  learning molecular fingerprints,'' in \emph{NIPS}, 2015, pp. 2224--2232.

\bibitem{chen2020physical}
J.~Chen, L.~Liu, H.~Wu, J.~Zhen, G.~Li, and L.~Lin, ``Physical-virtual
  collaboration graph network for station-level metro ridership prediction,''
  \emph{arXiv preprint arXiv:2001.04889}, 2020.

\end{thebibliography}

\begin{IEEEbiography}[{\includegraphics[width=1in,height=1.25in,clip,keepaspectratio]{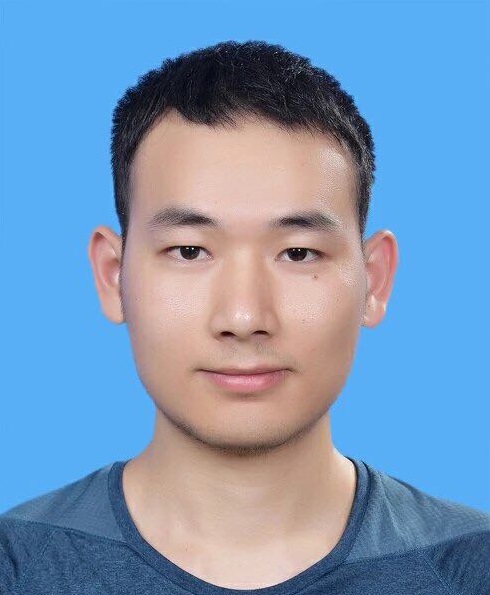}}]{Lingbo Liu}
received the B.E. degree from the School of Software, Sun Yat-sen University, Guangzhou, China, in 2015, where he is currently
pursuing the Ph.D degree in computer science with the School of Data and Computer Science. From March 2018 to May 2019, he was a research assistant at the University of Sydney, Australia. His current research interests include machine learning and intelligent transportation systems. He has authorized and co-authorized on more than 10 papers in top-tier academic journals and conferences.
\end{IEEEbiography}

\vspace{-10mm}
\begin{IEEEbiography}[{\includegraphics[width=1in,height=1.25in,clip,keepaspectratio]{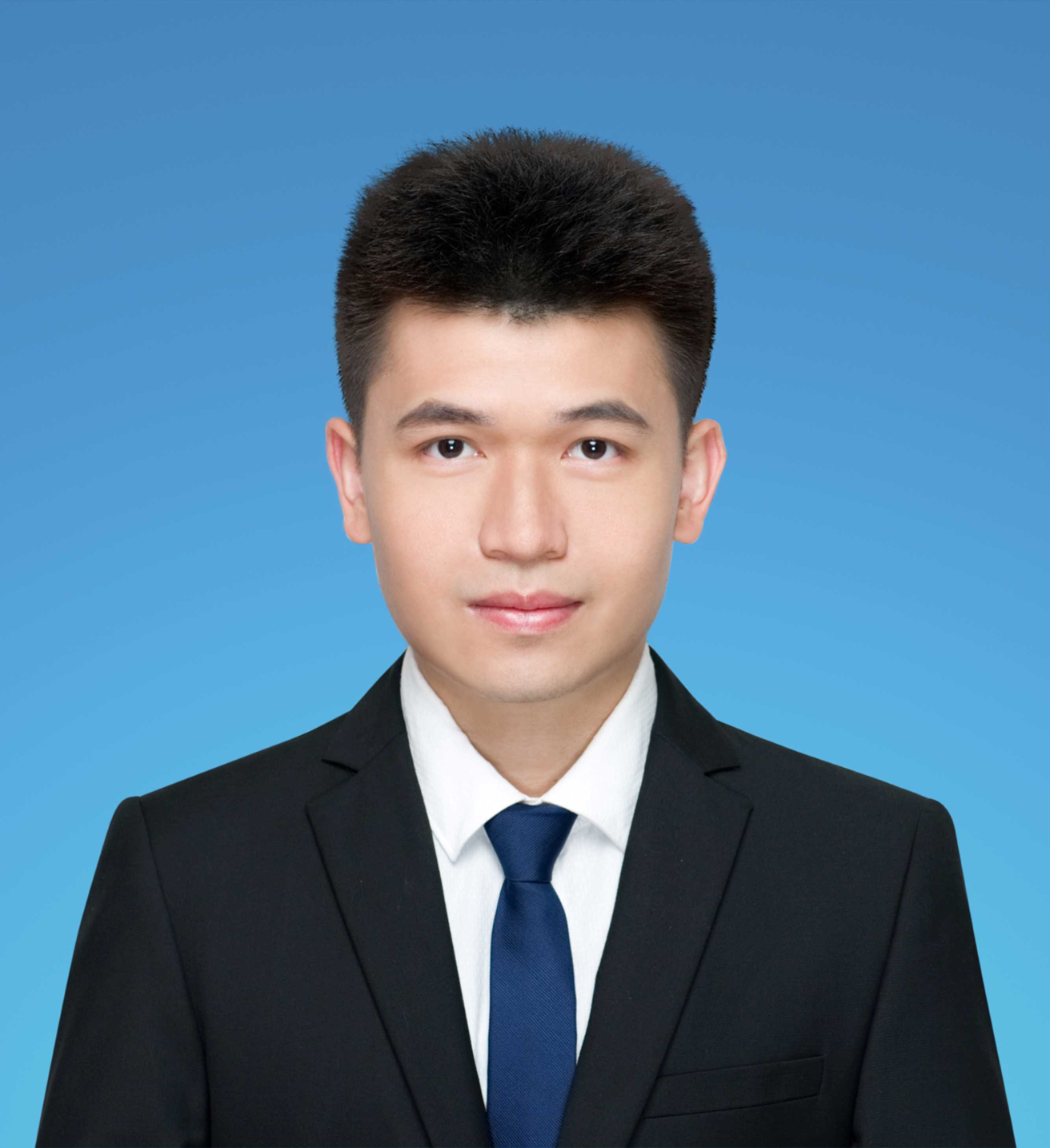}}]{Jiajie Zhen}
received the B.E. degree from the School of Mathematics, Sun Yat-sen University, Guangzhou, China, in 2018, where he is currently pursuing the Master's degree in computer science with the School of Data and Computer Science. His current research interests include computer vision and intelligent transportation systems.
\end{IEEEbiography}

\vspace{-10mm}
\begin{IEEEbiography}[{\includegraphics[width=1in,height=1.25in,clip,keepaspectratio]{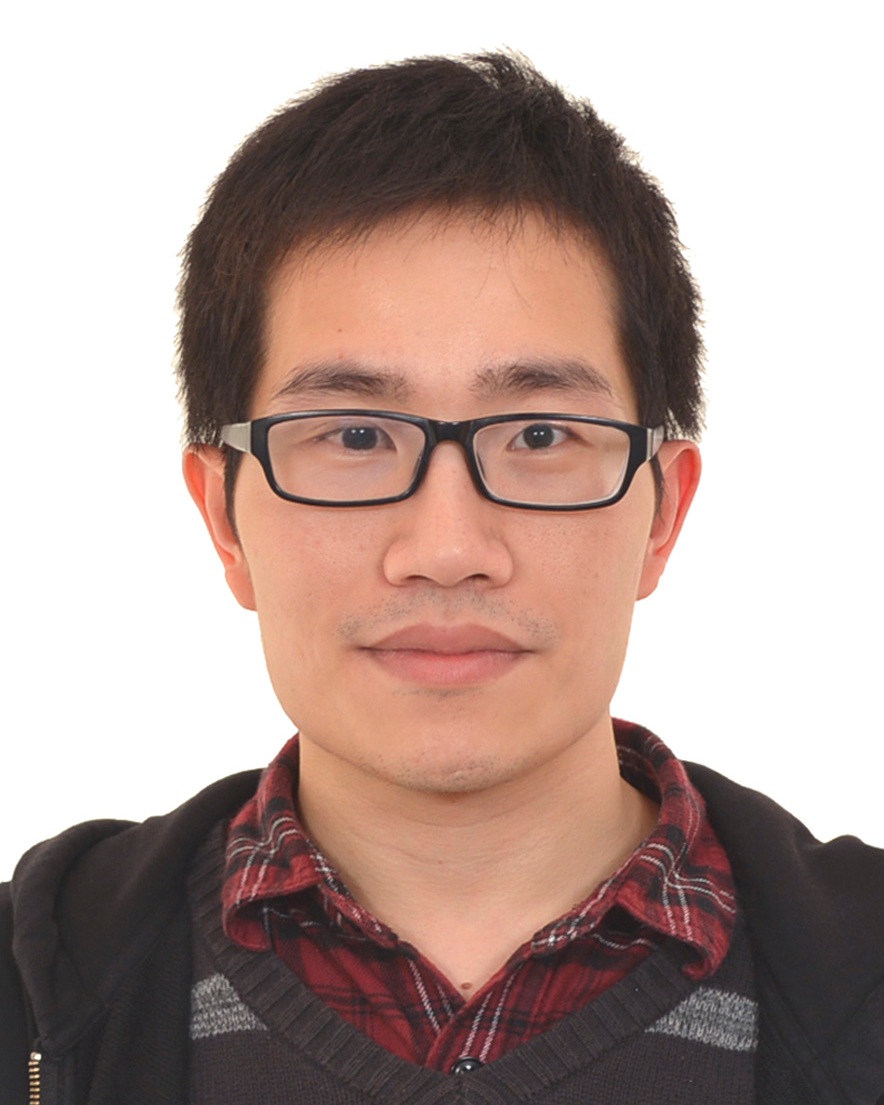}}]{Guanbin Li}
is currently a research associate professor in School of Data and Computer Science, Sun Yat-sen University. He received his PhD degree from the University of Hong Kong in 2016. He was a recipient of Hong Kong Postgraduate Fellowship. His current research interests include computer vision, image processing, and deep learning. He has authorized and co-authorized on more than 20 papers in top-tier academic journals and conferences.
He serves as an area chair for the conference of VISAPP. He has been serving as a reviewer for numerous academic journals and conferences such as TPAMI, TIP, TMM, TC, TNNLS, CVPR2018 and IJCAI2018.
\end{IEEEbiography}

\vspace{-10mm}
\begin{IEEEbiography}[{\includegraphics[width=1in,height=1.25in,clip,keepaspectratio]{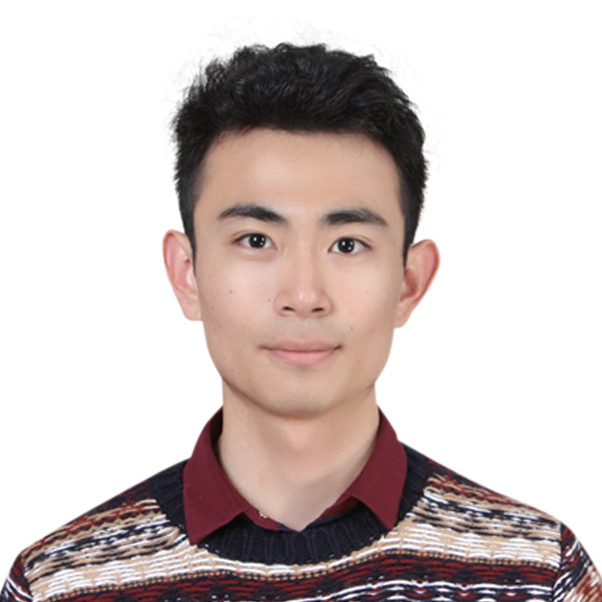}}]{Geng Zhan}
is currently pursuing the Master of Philosophy degree at the School of Electrical and Information Engineering, the University of Sydney. He received the B.E. degree from the Faculty of Electronic Information and Electrical Engineering, Dalian University of Technology in 2017. His current research interests include computer vision and machine learning.
\end{IEEEbiography}

\vspace{-10mm}
\begin{IEEEbiography}[{\includegraphics[width=1in,height=1.25in,clip,keepaspectratio]{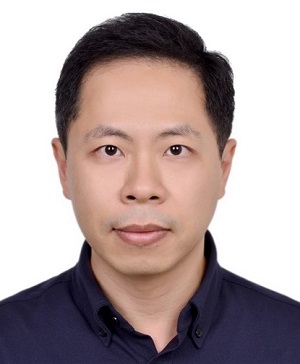}}]{Zhaocheng He}
received the B.S. and Ph.D. degrees from Sun Yat-Sen University, Guangzhou, China,
in 2000 and 2005, respectively. He is currently a Professor with the Guangdong Provincial Key Laboratory of Intelligent Transportation Systems (ITS), and the Research Center of ITS, Sun Yat-Sen University. His research interests include traffic flow dynamics and simulation, traffic data mining, and intelligent transportation systems.
\end{IEEEbiography}

\vspace{-10mm}
\begin{IEEEbiography}[{\includegraphics[width=1in,height=1.25in,clip,keepaspectratio]{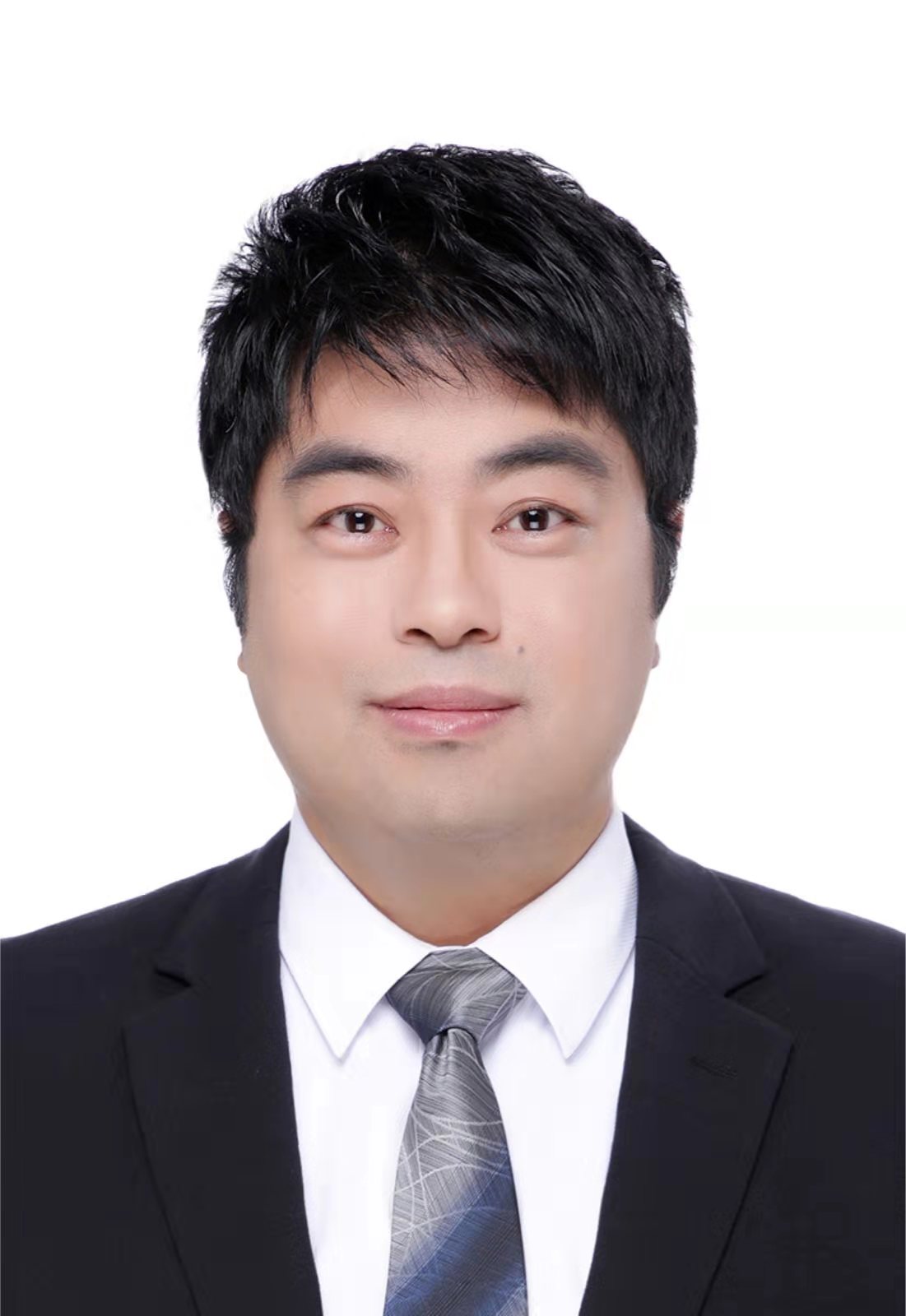}}]{Bowen Du}
received the Ph.D. degree in computer science and engineering from Beihang University, Beijing, China, in 2013. He is currently an professor with the State Key Laboratory of Software Development Environment, Beihang University. His research interests include smart city technology, multi-source data fusion, and traffic data mining.
\end{IEEEbiography}

\vspace{-10mm}
\begin{IEEEbiography}[{\includegraphics[width=1in,height=1.25in,clip,keepaspectratio]{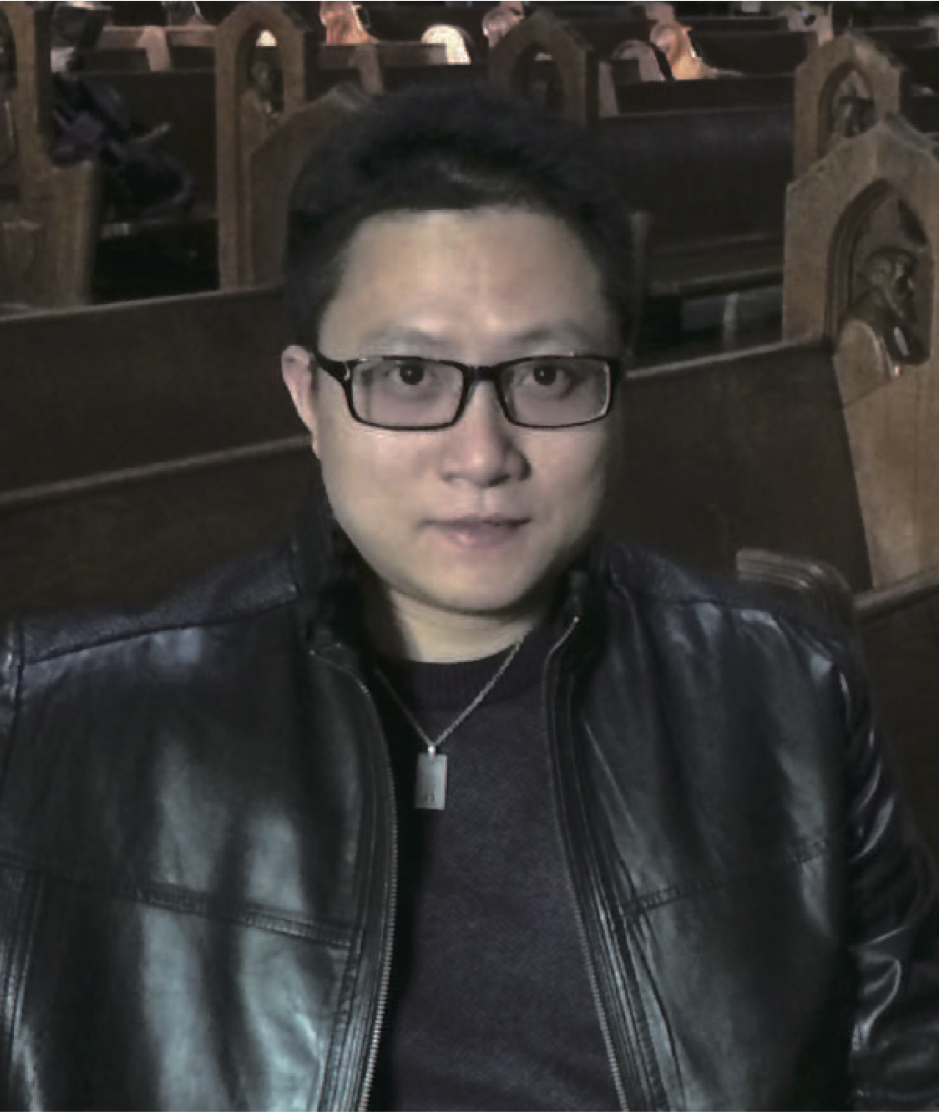}}]{Liang Lin}
is a full Professor of Sun Yat-sen University. He is the Excellent Young Scientist of the National Natural Science Foundation of China. From 2008 to 2010, he was a Post-Doctoral Fellow at University of California, Los Angeles. From 2014 to 2015, as a senior visiting scholar, he was with The Hong Kong Polytechnic University and The Chinese University of Hong Kong. He currently leads the SenseTime R\&D teams to develop cutting-edges and deliverable solutions on computer vision, data analysis and mining, and intelligent robotic systems. He has authorized and co-authorized on more than 100 papers in top-tier academic journals and conferences. He has been serving as an associate editor of IEEE Trans. Human-Machine Systems, The Visual Computer and Neurocomputing. He served as Area/Session Chairs for numerous conferences such as ICME, ACCV, ICMR. He was the recipient of Best Paper Runners-Up Award in ACM NPAR 2010, Google Faculty Award in 2012, Best Paper Diamond Award in IEEE ICME 2017, and Hong Kong Scholars Award in 2014. He is a Fellow of IET.
\end{IEEEbiography}




\end{document}